\documentclass{article}

\usepackage[preprint]{neurips_2025}  

\usepackage[utf8]{inputenc} 
\usepackage[T1]{fontenc}    
\usepackage{mathtools}
\usepackage{url}            
\usepackage{nicefrac}       
\usepackage{microtype}      
\usepackage[pdftex]{graphicx}
\usepackage{enumitem}

\usepackage{amsfonts}				
\usepackage{amsmath}				
\usepackage{siunitx}  				
\usepackage{xfrac}					
\usepackage{bm}						
\usepackage{mathtools}
\usepackage{mathabx}				
\usepackage{arydshln}  				
\usepackage{stmaryrd}  				
\usepackage{bbm}                    

\usepackage{booktabs}               
\usepackage{multirow}               
\usepackage{changepage,threeparttable} 
\usepackage{tabularx}               
\usepackage{etoolbox}               
\usepackage[dvipsnames]{xcolor}  
\definecolor{mycolor}{rgb}{0.5, 0.2, 0.8}
\definecolor{palepink}{rgb}{0.996, 0.753, 0.796}
\definecolor{mintyfresh}{rgb}{0.694, 0.878, 0.902}
\definecolor{lavenderdream}{rgb}{0.8, 0.6, 1}
\definecolor{powderblue}{rgb}{0.678, 0.847, 0.902}
\definecolor{peachykeen}{rgb}{1, 0.855, 0.725}
\definecolor{lightcoral}{rgb}{0.941, 0.502, 0.502}
\definecolor{seafoamgreen}{rgb}{0.627, 1, 1}
\definecolor{dustylavender}{rgb}{0.905, 0.753, 1}
\definecolor{softsage}{rgb}{0.702, 0.851, 0.824}
\definecolor{warmbeige}{rgb}{0.961, 0.863, 0.753}
\definecolor{darkelectricblue}{rgb}{0.33, 0.41, 0.47}
\definecolor{darkmidnightblue}{rgb}{0.0, 0.2, 0.4}

\usepackage[%
  colorlinks = true,
  citecolor  = darkmidnightblue,
  linkcolor  = darkmidnightblue,
  urlcolor   = darkmidnightblue,
  unicode,
  linktocpage,
]{hyperref}
\usepackage[capitalise,nameinlink]{cleveref}
\crefname{algocf}{alg.}{algs.}
\Crefname{algocf}{Alg.}{Algs.}
\Crefname{equation}{Eq.}{Eqs.}

\makeatletter
\AtBeginDocument
{
	\def\ltx@label#1{\cref@label{#1}}
	\def\label@in@display@noarg#1{\cref@old@label@in@display{#1}}
	\def\label@in@mmeasure@noarg#1{%
		\begingroup%
		\measuring@false%
		\cref@old@label@in@display{#1}
		\endgroup}%
} %
\makeatother

\usepackage{wrapfig}

\usepackage[disable]{todonotes}

\DeclareMathOperator*{\argmin}{arg\,min}					
\DeclareMathOperator{\diag}{diag}							
\DeclareMathOperator{\qr}{qr}
\DeclareMathOperator{\svd}{svd}
\DeclareMathOperator{\eigh}{eigh}
\DeclareMathOperator{\rank}{rank}								

\DeclareMathOperator{\norm}{norm}
\DeclareMathOperator{\vecc}{vec}

\newcommand\T{^{\top}} 										
\newcommand{\bigO}{\mathcal{O}}								

\newcommand{\R}{\mathbb{R}}					

\def\vomega{{\bm{\omega}}}

\def\vd{{\bm{d}}}
\def\ve{{\bm{e}}}
\def\vf{{\bm{f}}}

\def\vm{{\bm{m}}}

\def\vv{{\bm{v}}}

\def\vx{{\bm{x}}}

\def\mA{{\bm{A}}}
\def\mB{{\bm{B}}}
\def\mC{{\bm{C}}}
\def\mD{{\bm{D}}}

\def\mF{{\bm{F}}}

\def\mI{{\bm{I}}}
\def\mJ{{\bm{J}}}

\def\mL{{\bm{L}}}
\def\mM{{\bm{M}}}

\def\mP{{\bm{P}}}
\def\mQ{{\bm{Q}}}
\def\mR{{\bm{R}}}
\def\mS{{\bm{S}}}

\def\mU{{\bm{U}}}
\def\mV{{\bm{V}}}
\def\mW{{\bm{W}}}
\def\mX{{\bm{X}}}

\def\mZ{{\bm{Z}}}

\def\mPhi{{\bm{\Phi}}}
\def\mPsi{{\bm{\Psi}}}

\def\mSigma{{\bm{\Sigma}}}

\def\mOmega{{\bm{\Omega}}}
\def\mUpsilon{{\bm{\Upsilon}}}
\def\mGamma{{\bm{\Gamma}}}

\DeclareMathAlphabet{\mathsfit}{\encodingdefault}{\sfdefault}{m}{sl}
\SetMathAlphabet{\mathsfit}{bold}{\encodingdefault}{\sfdefault}{bx}{n}

\def\gL{{\mathcal{L}}}
\def\gM{{\mathcal{M}}}

\def\gP{{\mathcal{P}}}

\def\gR{{\mathcal{R}}}
\def\gS{{\mathcal{S}}}

\def\sC{{\mathbb{C}}}

\newcommand{\bigdim}{N}
\newcommand{\smalldim}{k}
\newcommand{\measdim}{p}

\newcommand{\defEq}{\coloneqq}  
\newcommand{\eqDef}{\eqqcolon}  

\newcommand{\ones}{\bm{1}}

\newcommand{\lowrankApprox}[1]{\llbracket #1  \rrbracket }

\newcommand{\frob}{\text{F}}

\newcommand{\measmat}{\mOmega}
\newcommand{\measmatBar}{\bar{\mOmega}}

\usepackage{acronym}
\acrodef{DL}[DL]{Deep Learning}
\acrodef{DNN}[DNN]{Deep Neural Network}
\acrodef{CNN}[CNN]{Convolutional Neural Network}
\acrodef{ResNet}[ResNet]{Residual Neural Network}
\acrodef{BN}[BN]{Batch Normalization}
\acrodef{MFLOPS}[MFLOPS]{Million of Floating Point Operations per Second}
\acrodef{LoRD}[LoRD]{Low-Rank plus Diagonal}
\acrodef{LoR}[LoR]{Low-Rank}
\acrodef{D}[D]{Diagonal}
\acrodef{MVP}[MVP]{Matrix-Vector Product}
\acrodef{ADMM}[ADMM]{Alternating Direction Method of Multipliers}
\acrodef{SGD}[SGD]{Stochastic Gradient Descent}
\acrodef{HVP}[HVP]{Hessian-Vector Product}
\acrodef{SVD}[SVD]{Singular Value Decomposition}
\acrodef{SSVD}[SSVD]{Sketched Singular Value Decomposition}
\acrodef{NTK}[NTK]{Neural Tangent Kernel}

\newcommand*{\eg}{e.g.\@\xspace}
\newcommand*{\ie}{i.e.\@\xspace}

\newcommand*{\wrt}{w.r.t.\@\xspace}
\newcommand*{\st}{s.t.\@\xspace}
\newcommand*{\versus}{vs.\@\xspace}
\newcommand{\sketchlord}{\textsc{Sketchlord}\@\xspace}
\newcommand{\xdiag}{\textsc{XDiag}\@\xspace}
\newcommand{\ssvd}{\textsc{SSVD}\@\xspace}

\usepackage[super]{nth}						                

\makeatletter
\newcommand{\oset}[3][0ex]{%
	\mathrel{\mathop{#3}\limits^{
			\vbox to#1{\kern-2\ex@
				\hbox{$\scriptstyle#2$}\vss}}}}
\makeatother

\newcommand\tightDots{\makebox[1em][c]{.\hfil.\hfil.}}

\makeatletter
\DeclareRobustCommand{\pdot}{\mathbin{\mathpalette\pdot@\relax}}
\newcommand{\pdot@}[2]{%
	\ooalign{%
		$\m@th#1\circ$\cr
		\hidewidth$\m@th#1\cdot$\hidewidth\cr
	}%
}
\makeatother

\usepackage{pifont}

\newcommand{\AF}[1]{\todo[color=palepink]{\textbf{AF:} #1}}
\newcommand{\FD}[1]{\todo[color=mintyfresh]{\textbf{FD:} #1}}
\newcommand{\FS}[1]{\todo[color=lavenderdream]{\textbf{FS:} #1}}

\usepackage[linesnumbered,ruled]{algorithm2e}  

\SetCommentSty{algoCommentFont}  

\makeatletter
\patchcmd{\@algocf@start}
  {-1.5em}
  {0pt}
  {}{}
\makeatother

\title{Sketching Low-Rank Plus Diagonal Matrices}

\author{%
  Andres Fernandez\\
  Tübingen AI Center, University of Tübingen\\
  \texttt{a.fernandez@uni-tuebingen.de} \\
  \And
  Felix Dangel\\
  Vector Institute, Toronto\\
  \texttt{fdangel@vectorinstitute.ai} \\
  \And
  Philipp Hennig\\
  Tübingen AI Center, University of Tübingen\\
  \texttt{philipp.hennig@uni-tuebingen.de} \\
  \And
  Frank Schneider\\
  Tübingen AI Center, University of Tübingen\\
  \texttt{f.schneider@uni-tuebingen.de} \\
}

\begin{document}
\maketitle

\begin{abstract}
Many relevant machine learning and scientific computing tasks involve high-dimensional linear operators accessible only via costly matrix-vector products.
In this context, recent advances in sketched methods have enabled the construction of \emph{either} low-rank \emph{or} diagonal approximations from few matrix-vector products.
This provides great speedup and scalability, but approximation errors arise due to the assumed simpler structure.
This work introduces \sketchlord, a method that simultaneously estimates both low-rank \emph{and} diagonal components,
targeting the broader class of Low-Rank \emph{plus} Diagonal (\acs{LoRD}) linear operators.
We demonstrate theoretically and empirically that this joint estimation is superior also to any sequential variant (diagonal-then-low-rank or low-rank-then-diagonal).
Then, we cast \sketchlord as a convex optimization problem, leading to a scalable algorithm.
Comprehensive experiments on synthetic (approximate) \acf{LoRD} matrices confirm \sketchlord's performance in accurately recovering these structures.
This positions it as a valuable addition to the structured approximation toolkit, particularly when high-fidelity approximations are desired for large-scale operators, such as the deep learning Hessian.
\end{abstract}

\section{Introduction}
\acresetall  
\label{sec:introduction}

High-dimensional linear operators are ubiquitous in modern data science, from the Hessian matrices of deep neural networks to large-scale scientific simulators.
Due to their sheer size, these operators are often accessible only \emph{implicitly} through their \acp{MVP}.
While \ac{MVP} access suffices for many applications, we sometimes prefer an \emph{explicit}, albeit possibly approximate, representation of the operator.
Such an explicit approximate representation can be useful, \eg, for visualizations or compressing an operator that contains simpler sub-structures. \FS{Not happy with this part. It needs better examples, a better description, and/or references.}
To achieve this tractably, one typically resorts to structured approximations
that are cheap to store and invert.

Recent advances in sketched methods enable the efficient construction of \emph{either} \acf{LoR} \emph{or} \acf{D} approximations from a comparably small number of \acp{MVP}.
However, these approaches can falter when the underlying linear operator exhibits a joint \ac{LoRD} structure, or can be well-approximated as such.
Attempting to recover operators with such inherent or approximate \ac{LoRD} structure using methods designed solely for either \ac{LoR} or diagonal components leads to suboptimal results, even when applying them sequentially (\cref{fig:toy,sec:toy}).
\FS{Alternative: Relying on methods for purely \ac{LoR} or diagonal structures, or even their sequential application, is inadequate for such composite operators, leading to demonstrably suboptimal recoveries (\cref{fig:toy,sec:toy}).}


To address this gap, we introduce \sketchlord, a novel sketched method designed for the \emph{joint} recovery of both \acf{LoR} \emph{and} \acf{D} components simultaneously from a few \acp{MVP} of large-scale linear operators.
\textbf{Our contributions are:}
\begin{itemize}[leftmargin=2em]
 \item We analytically show that for \ac{LoRD} operators, approximations for either \acf{LoR} or \acf{D} components individually, or even sequentially (\ac{LoR}-then-\ac{D} or \ac{D}-then-\ac{LoR}), are inherently suboptimal (\cref{sec:toy}). We empirically verify this for sketched variants.
 \item As a solution, we propose \sketchlord, a joint approximation method that phrases the \ac{LoRD} recovery as a convex optimization task, solvable efficiently via \acs{ADMM} (\cref{sec:sketchlord}). We provide practical accelerations, such as a new \emph{compact recovery} strategy for any sketched methods.\FS{Maybe add URL to code?}
 \item On synthetic matrices, we empirically show that \sketchlord achieves high-quality approximations across a broad class of (approximately) \ac{LoRD} operators (\cref{sec:synth}). We also observe that \sketchlord is stable for a wide range of its hyperparameters (\cref{sec:sketchlord_algo}).
\end{itemize}

\section{Background}
\label{sec:background}

Modern scientific computing and machine learning frequently encounter high-dimensional matrices, such as discretized differential operators in physical simulations\FS{Citation?} or large covariance matrices\FS{Citation?}.
A prominent example motivating this work is the Hessian matrix in deep learning.
Fundamental operations, like storing ($\bigO(\bigdim^2)$), performing explicit \acp{MVP} ($\bigO(\bigdim^2)$), or computing its inverse ($\bigO(\bigdim^3)$), become infeasible even for comparably modest neural networks parameter counts, $\bigdim$, in the millions.
Implicit, matrix-free, \acp{MVP}---\eg, via Pearlmutter's trick for the Hessian \citep{fast_hvp}---and structured approximations offer ways to nevertheless work with these massive matrices in practice.
This work concentrates on achieving high-fidelity approximations for operators that are inherently \acf{LoRD} or can be accurately represented by such a structure.
\FS{I don't know whether this is the right place for this statement, but I feel like we need to make it somewhere.}

\textbf{Structured approximations.}
To render large operators tractable, various structured approximations are employed in practice.
Low-rank captures dominant modes while diagonal approximations simplify interactions to element-wise scaling.
Another instance, K-FAC \citep{Martens2015}, a popular Hessian approximation, employs a Kronecker-factored block-diagonal structure, primarily for computational efficiency in optimization.
Crucially, the ``best'' approximation strategy is determined not only by the operator's intrinsic structure but also by a critical trade-off:
the application's demand for \emph{approximation quality} (low approximation error) versus constraints on the acceptable \emph{runtime} and available \emph{memory}.
In essence, the specific application dictates whether a fast-but-coarse approximation suffices or if a potentially slower but more accurate representation is desired, and how memory constraints influence this choice.
\FS{I don't know whether here is the right place to introduce this point, but I feel this is important. Crucially, it let's us explain why \sketchlord is useful, even if its runtime is quite large: It offers an additional tool in the practitioner's toolbox and might be the right tool, depending on this trade-off. We can refer back to this point when transition from the limitations to the conclusion, \ie ``yes, it is computationally expensive, but it still might be the right tool in certain situations''.}
In this paper, we focus on the composite \acf{LoRD} structure.
Evidence suggests that important and complex operators, such as the Hessian matrix in deep learning, may be well-approximated by such a structure.
For example, spectral analyses show dominant low-rank components \citep[\eg][]{hessian_bulk,papyan18_spectrum,large_scale_spectrum}, while common practices like diagonal regularizers introducing a diagonal contribution.
Developing robust methods for high-quality \ac{LoRD} approximation is thus of significant practical interest.
\FS{For the moment, I cut the section on Hessians, since it is not really used. I added two sentences to this paragraph here, that LoRD structures might be useful, since important matrices like the Hessian might be well-approximated with such a structure.}


\textbf{Sketched methods for structured recovery.}
Sketched methods, based on random matrix theory, offer a data-efficient paradigm for constructing these structured approximations from only a limited number, $p$, of \ac{MVP} measurements.
Due to their scalability, reliability, and ability to be parallelized \citep{rmt_svd}, they have steadily gained popularity in many applications including signal processing \citep{rla_sigproc}, scientific computing \citep{turnstile,ssvd19}, or machine learning \citep{rla_ml,fernandez}.
Here, we focus on sketched methods for low-rank and diagonal recovery.
Sketched methods like \xdiag \citep{xtrace} or \textsc{Diag++} are based on the Girard-Hutchinson estimator\FS{Citation?} and allow estimating $\diag(\mA)$ from only $p \!\ll\!\smalldim \eqDef\rank(\mA)$ random measurements.
In this work, we compare to \xdiag which exhibits improved measurement-efficiency.
Similarly, a \ac{SSVD} provides an approximate \acs{SVD} of a low-rank operator, using $p$ left and right measurements, followed by QR orthogonalization, and an \acs{SVD} of size $p \!\times\! p$ (see \Cref{alg:ssvd}).
Crucially, using $p$ slightly larger than $\smalldim$ is enough with high probability \citep{rmt_svd,tropp_svd17}.
When $\mA$ is not truly low-rank, an oversampled recovery mechanism can be used instead to reduce noise \citep{boutsidis,ssvd19}

\textbf{Joint recovery with rank minimization.}\FS{My radical proposal would be to move this entire paragraph to the appendix (\eg ``Detailed mathematical derivation'') and remove it from the background section.}
Consider the following general problem of rank minimization under linear mapping $\gM$ as \eg studied in \citet{nuclear_convex}:
\begin{align}
(P_0) \qquad \mX_\natural = \argmin_{\mX} ~~\rank(\mX)  ~~~~~\text{\st}~~~~~ \gM(\mX) = \mM
\end{align}
Though NP-hard in general \citep{rank_nphard}, its convex relaxation is equivalent with high probability when $\gM$ is sufficiently random and yields enough measurements \citep{nuclear_convex}:
\begin{align}
(P_1 \equiv P_0) \qquad \mX_\natural = \argmin_{\mX} ~~\lVert \mX \rVert_*  ~~~~~\text{\st}~~~~~ \gM(\mX) = \mM
\end{align}
This objective, in turn, admits a Lagrangian characterization that is equivalent for some $\lambda\!\in\!\R_{\geq 0}$:
\footnote{Analytical $\lambda$ is unavailable, with a few exceptions \citep[\eg][]{rpca,closed_form_lambda}.}
\begin{align}
(P_\lambda \equiv P_1) \qquad \mX_{\natural} =  \argmin_{\mX}~~  \frac{1}{2} \underbrace{ \lVert \mM - \gM(\mX) \rVert_\frob^2}_{\gL_2}  ~~+~~  \lambda  \lVert \mX \rVert_* 
\end{align}
\\[-0.5em]
To solve it efficiently, \citet{sv_thresh_algo} proposed to use an instance of \acs{ADMM} \citep{admm}, that in its simplest form, runs the following \emph{projected gradient} step iteratively (for step size $\eta \in \R_{> 0}$):
\begin{align}
  \label{eq:projgrad}
   \mX \leftarrow \gP_\lambda(\mX - \eta \nabla_{\mX} \gL_2)
\end{align}
 where $\gP_\lambda$ is the \emph{spectral shrinkage} operator, computed by applying a soft threshold to the singular values of its input (optimality is discussed in \citep[Th.2.1]{sv_thresh_algo}).

\section{The need for joint recovery}
\label{sec:toy}

\begin{figure}[t!]  
	\centering
	\includegraphics[width=0.97\textwidth]{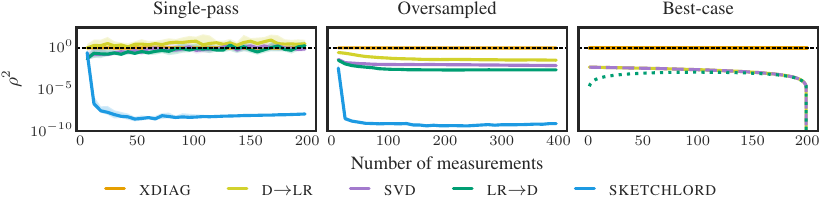}
	\caption{
		\textbf{Joint recovery is superior for the \acf{LoRD} operator $\mA \!=\! \ones \ones\T  \!+\! \mI$:}
		\textit{(Left \& Center)} Empirical recovery performance of various sketched \acf{LoR} and \acf{D} approximation methods versus number of measurements, using single-pass or oversampled recovery.
		\sketchlord's joint recovery of the \ac{LoR} and \ac{D} component consistently yields superior approximations compared to individual or sequential recovery strategies.
		Medians (thick lines) and interquartile ranges (shaded region, 25th-75th percentiles, $30$ samples) are shown.
		For reference, the dashed line marks $100\%$ relative error ($\rho^2$), below which methods outperform an all-zero recovery.
		\textit{(Right)} Theoretical best-case recovery error bounds for different \ac{LoR}/\ac{D} recoveries, as derived in \cref{sec:toy}.
		\sketchlord is omitted due to its zero theoretical error.
	}
	\label{fig:toy}
\end{figure}
\FS{We should make the legend and main text more consistent. My preference would be to use ``Diagonal (D)'', ``Low-Rank (LoR)'', ``$\text{LoR} \!\rightarrow\! \text{D}$'', and ``$\text{D} \!\rightarrow\! \text{LoR}$''. The one in parenthesis would be the short forms.}

To illustrate the necessity for \emph{joint} \acf{LoRD} recovery, we consider a canonical instance of this operator class, $\mA \defEq \ones \ones\T  \!+\! \mI \!\in\! \R^{\bigdim \times \bigdim}$ with $\bigdim \!\gg\! 1$.
This operator, while simple, encapsulates the core challenge: its structure is an explicit sum of a rank-$1$ component ($\ones \ones\T$) and a full-rank diagonal component ($\mI$).
In this section, we first demonstrate that attempting to recover this structure through separate or sequential
\FS{I wonder whether ``sequential'' is a good term, since in some sense \sketchlord also does sequential approximations, just iteratively, no? \textbf{AF:} sketchlord is sequential due to ADMM, but it does optimize the joint objective, so it is joint in nature}
approximations is fundamentally suboptimal (\cref{sec:toy_theory}).
After that, we empirically verify these findings (\cref{sec:toy_experiment}) by applying sketched versions of these strategies to $\mA$ (with $N\!=\!200$).
To measure the recovery error of an approximation $\hat{\mX}$ for a linear operator $\mX$, we use the \emph{residual energy} (lower is better) defined as:
\begin{align}
  \label{eq:rho}
   \rho^2_{\hat{\mX}}\big(\mX \big) \defEq  \frac{\lVert \mX - \hat{\mX} \rVert_\frob^2}{\lVert \mX \rVert_\frob^2} \quad \in \R_{\geq 0} .
\end{align}

\subsection{Theoretical analysis}
\label{sec:toy_theory}


\textbf{Diagonal approximation.} Approximating $\mA$ by its diagonal, $\hat{\mA} \!=\! \diag(\mA) \!=\! 2\mI$, 
the error is:
\begin{align}
   \rho^2_{\text{D}} \big(\mA\big) = \frac{\lVert \mA - \diag(\mA) \rVert_F^2}{\lVert \mA \rVert_F^2} &= \frac{\bigdim^2 - \bigdim }{\bigdim^2 + 3 \bigdim} = \frac{\bigdim - 1}{\bigdim + 3} \,,
\end{align}
\ie \emph{purely diagonal approximations reach near worst-case error already for moderate $\bigdim$}.

\textbf{Low-rank approximation.} To analize the rank-$\smalldim$ optimal approximation: $\hat{\mA} \!=\! \lowrankApprox{\mA}_{\smalldim} = \sum_{i=0}^{\smalldim - 1} \lambda_i \vv_i \vv_i\T$, with eigenvectors $\vv_i$ and eigenvalues $\lambda_i$, we observe that $\mA$ is a Toeplitz matrix, and thus admits the following eigendecomposition:
\FD{I would remove $\cdot$s and start counting from 1 rather than 0. This is more common in the literature imo. \textbf{AF:} I see that, $\vf_0$ shows up often in sigproc when talking about the zero-frequency, as a proxy for $\vf_\omega$}
$\mA \!=\! \mF^{-1} (\bigdim\!\cdot\!\ve_0 \ve_0\T \!+\! \mI) \mF$,
where $\mF$ is the discrete Fourier basis of corresponding dimension and $\ve_i$ is the standard basis vector with nonzero value at the (zero-indexed) $i$-th position.
We can then express the $\rho$-optimal rank-$\smalldim$ approximation analytically as: 
\FD{Is this standard notation? I have never seen it in this context before. \textbf{AF:} yes, see e.g. \cite{ssvd19}}
$\lowrankApprox{\mA}_{\smalldim} \defEq \bigdim \cdot \vf_0 \vf_0^* +  \sum_{i=0}^{\smalldim - 1} \vf_i \vf_i^*$
and the error for the resulting optimal rank-$\smalldim$ deflation becomes
\begin{align*}
   \rho^2_{\text{LoR}} \big(\mA\big) &= \frac{\bigdim - \smalldim}{\bigdim^2\!+\!3\bigdim} = \frac{1 - \sfrac{\smalldim}{\bigdim}}{\bigdim + 3} \quad \in \Big[0,~~ \frac{1 - \sfrac{1}{\bigdim}}{\bigdim + 3} \Big]\,.
\end{align*}
This is drastically better than a diagonal approximation due to the dominance of the $\vf_0$ component.
For an operator of the shape $\mI + \sigma \ones \ones\T$ and $\sigma \!\ll\! 1$, this approximation would not be that favorable.
Furthermore, this approximation entails a poor tradeoff for all $\smalldim \geq 2$, since each additional eigenpair only improves the error by a small fraction, and a zero-error approximation requires $\smalldim \!=\! \bigdim$.

\textbf{Diagonal-then-low-rank.} This sequential approximation, first extracts the diagonal and then applies a low-rank approximation to the residual, \ie $\hat{\mA} \!=\! \diag(\mA) \!+\! \lowrankApprox{\mA \!-\! \diag(\mA)}_{\smalldim}$.
The residual, $\bar{\mA} \!\defEq\! \mA - \diag(\mA) \!=\! \mF^{-1} (\bigdim \cdot \ve_0 \ve_0\T - \mI) \mF$ remains full-rank for $\bigdim \!\geq\! 2$ making the \emph{error identical to low-rank approximation}, without any benefits from the additional computation:
\begin{align*}
    \rho^2_{\text{D} \rightarrow \text{LoR}} \big(\mA \big)
    = \frac{\lVert \mA - (\diag(\mA) \!+\! \lowrankApprox{\bar{\mA}}_\smalldim) \rVert_\frob^2}{\lVert \mA \rVert_\frob^2} 
    = \frac{\lVert \bar{\mA} \!-\! \lowrankApprox{\bar{\mA}}_\smalldim \rVert_\frob^2}{\lVert \mA \rVert_\frob^2} 
    = \frac{\bigdim - \smalldim}{\bigdim^2 + 3 \bigdim} 
    = \rho^2_{\text{LoR}} \big(\mA\big) \,.
\end{align*}

\textbf{Low-rank-then-diagonal.} We reverse the order, starting with low-rank deflation followed by diagonal estimation of the residual, \ie $\hat{\mA} \!=\! \lowrankApprox{\mA}_\smalldim \!+\! \diag(\mA \!-\! \lowrankApprox{\mA}_\smalldim)$.
The residual has the form $\mA \!-\! \lowrankApprox{\mA}_{\smalldim} \!=\! \sum_{i=\smalldim}^{\bigdim - 1} \vf_i \vf_i^*$ ($\smalldim \geq 1$), and we get the following recovery error (see \cref{app:toy_theory}):
\begin{align*}
    \rho^2_{\text{LoR} \rightarrow \text{D}} \big(\mA \big)
    &= \frac{\lVert \mA - \lowrankApprox{\mA}_\smalldim + \diag(\mA - \lowrankApprox{\mA}_\smalldim)  \rVert_\frob^2}{\lVert \mA \rVert_\frob^2}
    = \frac{\sfrac{\smalldim}{\bigdim} - \sfrac{\smalldim^2}{\bigdim^2}}{\bigdim + 3}
\end{align*}
While strictly better than $\rho^2_{\text{LoR}}$ (for $\smalldim \!<\!\bigdim$), the error does not decrease monotonically as $\smalldim$ increases, making the optimal trade-off between low-rank deflation and diagonal estimation non-trivial.
\FD{Add a statement on what the trade-off is, \ie, why sometimes the error may increase although we use a higher rank.}

These bounds (\cref{fig:toy}, right) underscore the inherent limitations of non-joint methods.
Crucially, a perfect joint \ac{LoRD} recovery, would achieve zero error for this example operator.

\subsection{Empirical verification for sketched methods}
\label{sec:toy_experiment}

We now empirically extend our findings for sketched variants of the strategies described in \cref{sec:toy_theory}.
The goal is to evaluate their practical performance when limited to a small number of \ac{MVP} measurements, which we consider to be the primary bottleneck in this scenario.
We thus prioritize measurement efficiency in our baseline to ensure a fair comparison.
Specifically, we estimate diagonal components using \xdiag (\citet{xtrace}, \Cref{alg:xdiag}) and low-rank components via an \acs{SSVD} (\citet{tropp_svd17}, \Cref{alg:ssvd}).
These also form the building blocks for the sequential strategies.
$\text{D} \!\rightarrow\! \text{LoR}$ (\Cref{alg:diag_then_lowrank}) first estimates the diagonal via \xdiag and then uses the \ssvd on the residual.
Since \xdiag performs an initial low-rank deflation, as part of its internal mechanism, we can reuse these measurements also for the \ssvd, making this baseline more measurement efficient than a naive implementation.
Conversely in $\text{LoR} \!\rightarrow\! \text{D}$ (\Cref{alg:lowrank_then_diag}), we already perform a low-rank deflation via the \ssvd, and can thus bypass the initial deflation step in the \xdiag method and directly apply the Girard-Hutchinson method, reminiscent of methods like \textsc{diag++}\FS{Citation?}.
This, again, optimizes the baseline in terms of measurement efficiency, providing a simple yet fair comparison.


\Cref{fig:toy} contrasts theoretical error bounds (right panel) with the empirical performance of sketched algorithms (left: single-pass, center: oversampled recovery strategy) for the model operator $\mA \in \R^{200 \times 200}$.
\footnote{While the center and left panels show the empirical performance \versus the number of measurements $p$, the right panel shows the theoretical error bound \versus the rank $\smalldim$ of an ideal recovery. However, since $p\!=\!\bigO(\smalldim)$, this serves as a useful estimate also \versus the number of measurements $p$.}
\FS{@Andres, reminder to add something to the discussion about the different x-axis in \cref{fig:toy}. I tried to address this with this footnote but it needs your expertise. I think $p>k$, but since the theoretical bound is not monotonously, it is not a true bound, even in this case.}
The empirical results reinforce the need for joint \ac{LoRD} recovery uncovered in our theoretical analysis.
As expected, all algorithms yield recovery errors equal or larger than the respective theoretical lower bound.
Individual approximation methods perform suboptimally, and the sequential methods offer only little improvement.
$\text{LoR} \!\rightarrow\! \text{D}$ is only marginally better than a low-rank only approach, while $\text{D} \!\rightarrow\! \text{LoR}$ performs worse here.
In stark contrast, a joint recovery approach (represented by \sketchlord, introduced in \Cref{sec:sketchlord}) achieves significantly lower error, often by many orders of magnitude, for the same number of measurements.
This empirical evidence strongly underlines the necessity of algorithms explicitly designed for joint \ac{LoRD} recovery when encountering operators with this composite structure.
\FS{I kept it relatively short when discussing the empirical results. There is more to draw from this plot though, see Andres' notes in the comments. Feel free to include this information here.}
\FS{I keep using ``recovery strategy'' or similar terms for both \xdiag/\ssvd/etc. and single-pass/oversampled which is confusing. Suggestions for better terms?}
\AF{Do we agree on an acronym for low-rank plus diagonal matrices? LoRD may be a bit too comic, maybe L+D? \textbf{FS:} I am fine with LoRD, since that explains why we call it \sketchlord. We could also do LoR+D, but I am also happy with L+D.}

\section{Sketched low-rank plus diagonal approximations}
\label{sec:sketchlord}

This section details our core contribution:
a scalable method for estimation of \acf{LoRD} operators.
We first introduce a sketched mechanism and constraint that allow us to express joint low-rank and diagonal estimation as a convex optimization problem (\cref{sec:sketchlord_convex}).
This formulation leads to our scalable strategy, called \sketchlord (\Cref{alg:sketchlord}), which we refine with several accelerations critical for practical efficiency (\cref{sec:sketchlord_algo}).


\subsection{A convex optimization strategy}
\label{sec:sketchlord_convex}

Consider a square\footnote{The extension to non-square operators is supported in our code but omitted here for clarity.} 
linear operator $\mA \!\in\! \sC^{\bigdim \!\times\! \bigdim}$ that is only accessible via (costly) \ac{MVP} measurements, $\vm \!=\! \mA \vomega$, but admits a low-rank plus diagonal decomposition $\mA \!\eqDef\! \mL \!+\! \mD$, for $\rank(\mL) \!=\! \smalldim \ll \bigdim$ and $\mD \!\defEq\! \diag(\vd)$.
Since this decomposition is not necessarily unique (consider \eg scalar matrices), we are interested in a rank-minimal solution:
\begin{align}
(P_+) \qquad \mL_{\natural}, \vd_\natural = \argmin_{\mL, \vd} ~~\rank(\mL)  ~~~~~\text{\st}~~~~~ \mL = \mA + \diag(\vd)
\end{align}
This characterization is generally NP-hard \citep{rank_nphard} and the constraint involves unknown quantities $\mA$ and $\vd$.
To make it tractable, we will use a particular class of matrices $\measmat, \measmatBar \!\in\! \sC^{\bigdim \!\times\! \measdim}$ satisfying $\measmat_{i, j} \measmatBar_{i, j} \!=\! 1 \,\forall\, (i, j)$.
We will also assume that it is possible to recover $\mL$ from $\mX \!\defEq\! \mL \measmat$, which is true with high probability if $\mOmega$ is random and we have enough measurements $\measdim\!>\!\smalldim$ (\cref{sec:background}).
Then, the following sketching mechanism exposes an interesting structure:
\begin{align}
    \mM \defEq \mA \measmat \pdot \measmatBar = \mL \measmat \!\pdot \measmatBar + (\mD \measmat)\!\pdot \measmatBar \eqDef \mX \pdot \measmatBar + \vd \ones^*\,.
\end{align}
Namely, that $\vd \ones^* \!=\!\vd \ones^* \frac{1}{\measdim}(\ones \ones^*)$. This can be used to define a constraint that only depends on $\mX$:
\begin{align}
  \label{eq:equality_constraint}
\mM - \mX \pdot \measmatBar = (\mM - \mX \pdot \measmatBar) \frac{1}{\measdim}\ones \ones^*
\Rightarrow
\underbrace{\mM \big( \mI \!-\! \frac{1}{\measdim}\ones \ones^* \big)}_{\tilde{\mM}} =  (\mX \pdot \measmatBar) \big( \mI\!-\! \frac{1}{\measdim}\ones \ones^* \big)
\end{align}
\\[-0.5em]
This constraint defines a \emph{feasible set} of candidate solutions $\mX$, and it is easy to verify that $\mL_\natural$ is always contained in this set.
This, combined with the fact that we assumed sufficient measurements for recovery, \ie $\rank(\mL) \!=\! \rank(\mL \measmat)$, allows us to express our original problem in the following more tractable, re-parametrized form:
\begin{align}
(P_+' \equiv P_+) \qquad  \mX_{\natural} = \argmin_{\mX} ~~\rank(\mX)  ~~~~~\text{\st}~~~~~ (\mX \!\pdot\!\measmatBar) \big( \mI\!-\! \frac{1}{\measdim}\ones \ones^* \big) = \tilde{\mM}
\end{align}
This is now an instance of the canonical low-rank problem reviewed in \Cref{sec:background}, which, as discussed, can be relaxed to the following equivalent Lagrangian for some unknown $\lambda$:
\begin{align}
\label{eq:problem_lambda}
(P_\lambda \equiv P_+') \qquad \mX_{\natural} =  \argmin_{\mX}~~ \underbrace{ \frac{1}{2}  \lVert \tilde{\mM} - (\mX \!\pdot\!\measmatBar) \big( \mI\!-\! \frac{1}{\measdim}\ones \ones^* \big) \rVert_\frob^2}_{\gL_2}  ~~+~~  \lambda  \lVert \mX \rVert_*\,,
\end{align}
And this can be optimized iteratively via $\mX \leftarrow \gP_\lambda(\mX - \eta \nabla_{\mX} \gL_2)$ (\Cref{sec:background}), with the following gradient (see \Cref{app:convexity} for a derivation):
\begin{align}
  \label{eq:admm_gradient}
  \nabla_{\mX} \gL_2 = \Big[ \Big( (\mX \pdot \measmatBar) - \mM   \Big) \big( \mI\!-\! \frac{1}{\measdim}\ones \ones^* \big) \Big ] \pdot \measmatBar \,.
\end{align}
Yielding the optimal $\mX_\natural = \mL_\natural \measmat$.
Recall that we assumed a recovery of $\mL_\natural$ from $\mX_\natural$ is possible.
One important caveat here is that this requires adjoint measurements from $\mL_\natural$, the very operator we seek \citep[3.5]{tropp_svd17}.
We resolve this by first isolating the diagonal component, and then performing deflated measurements as follows:
\begin{align}
  \mD_\natural = \sfrac{1}{\measdim} \diag \big( (\mM - \mX_\natural \pdot \measmatBar ) \ones\big) \qquad \text{and then} \qquad
  \mUpsilon^* \mL_\natural = (\mUpsilon^* \mA) - (\mUpsilon^* \mD_\natural) \label{eq:diag_recovery_deflated_meas}
\end{align}
With this, we can recover $\mL_\natural$, finalizing the procedure.
Putting everything together, we obtain \textsc{sketchlord}, our full algorithm as presented in \Cref{alg:sketchlord}.
\AF{Mention somewhere that the recovery of $\mL$ is low-rank, as in the SSVD, so we remain memory-efficient which is the whole point}

\begin{minipage}[t]{0.48\textwidth} 
	{
\begin{algorithm}[H]
	\small
  \DontPrintSemicolon
  \KwIn{$\mA\!\in\!\sC^{\bigdim\!\times\!\bigdim}$ \tcp*{Linear operator}}
  \KwIn{$\mOmega, \mUpsilon \!\in\!\sC^{\bigdim\!\times\!\measdim}$   \hspace{-15pt} \tcp*{Random matrices}}
  \KwIn{$\eta, \lambda \!\in\!\R_{>0}$}
  \tcp{Forward random measurements}
  $\mM \leftarrow \mA \mOmega \pdot \bar{\mOmega}$ \hspace{-1cm} \tcp*{$\sC^{\bigdim\times \measdim}$}
  \tcp{Constrained rank minimization}
  $\mX \leftarrow \mM \pdot \mOmega$ \\
  \While{not converged}{
    $\mX \leftarrow \gP_\lambda(\mX - \eta \nabla_{\mX} \gL_2)$
  }
  \tcp{Diagonal recovery and adjoint measurements}
  $\mD \leftarrow \sfrac{1}{\measdim} \diag\big( (\mM - \mX) \ones \big)$ \hspace{-1cm}  \tcp*{$\sC^{\bigdim}$}
  $\mW^* \leftarrow (\mUpsilon^* \mA) - (\mUpsilon^* \mD)$ \hspace{-1cm} \tcp*{$\sC^{\measdim \times \bigdim}$}
  \tcp{Recovery}
  $(\mU, \mSigma, \mV^*) \leftarrow \textsc{compact}(\mM, \mOmega, \mW)$\\
  \Return{$(\mU, \mSigma, \mV^*), \vd$} \tcp*{$\mU \mSigma \mV^* + \diag(\vd) \approx \mA$}
  \caption{\textsc{sketchlord}}
  \label{alg:sketchlord}
\end{algorithm}
}

\end{minipage}
\hfill 
\begin{minipage}[t]{0.48\textwidth} 
	{
\begin{algorithm}[H]
	\small
  \DontPrintSemicolon
  \KwIn{$\mM, \mOmega, \mW \!\in\!\mathbb{C}^{\bigdim\!\times\!\measdim}$}
  $(\mP, \mS) \leftarrow \qr(\mM)$  \tcp*{$\sC^{\bigdim \times \measdim}$}
  $(\mQ, \mR) \leftarrow \qr(\mW)$  \tcp*{$\sC^{\bigdim \times \measdim}$}
  $\mB \leftarrow \mQ^* \mOmega$  \tcp*{$\sC^{\measdim \times \measdim}$}
  $\mPsi \leftarrow \mS \mB^\dagger$ \tcp*{$\sC^{\measdim \times \measdim}$}
  $\mZ, \mSigma^2 \leftarrow \eigh(\mPsi^* \mPsi)$ \tcp*{$\sC^{\measdim \times \measdim}$}
  $\mV \leftarrow \mQ \mZ$ \tcp*{$\sC^{\measdim \times \measdim}$}
  $\mU \leftarrow \mP (\mB \mS^\dagger)^* \mZ \mSigma$ \tcp*{$\sC^{\bigdim \times \measdim}$}
  \Return{$(\mU, \mSigma, \mV^*)$}
\caption{\textsc{compact}}
\label{alg:compact}
\end{algorithm}
}

\end{minipage}


\subsection{The \textsc{sketchlord} algorithm}
\label{sec:sketchlord_algo}

Overall, \sketchlord can be seen as an extension of \textsc{ssvd} (\Cref{alg:ssvd}) featuring additional rank-minimization and diagonal deflation steps between measurements and recovery.
But in reality, we incorporate a few modifications that make it practical, highlighted in \cref{fig:accelerations} and discussed below:

\begin{figure}[t]
  \centering
   \includegraphics[width=0.85\textwidth]{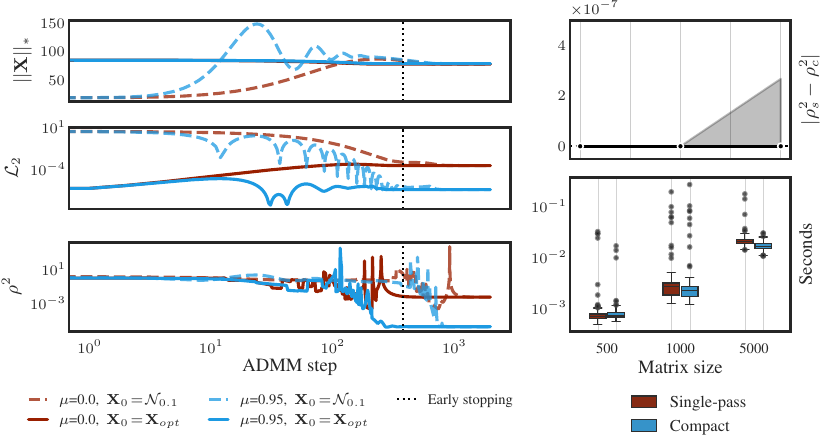}
   \caption{\textbf{\textsc{sketchlord} accelerations:}
     \emph{(Left)} Evolution of the $P_\lambda$ losses from \Cref{eq:problem_lambda} as a function of optimization step, together with the energy error metric $\rho$ defined in \Cref{eq:rho}. We see that gradient momentum (blue) provides faster and better convergence, and we also see that an optimal initialization (solid lines) helps with performance and stability, particularly in combination with momentum.
     We also see how our simple early stopping strategy (evaluated here on the run with momentum and optimal initialization) correctly and efficiently predicts convergence of $\rho$.
     \emph{(Bottom right)} Already for small scales, our proposed compact recovery provides a visible speedup, which is projected to improve as problems grow in size.
     \emph{(Top right)} The $\rho$ difference between our proposed \textsc{compact} recovery (\Cref{alg:compact}) and  \textsc{singlepass} (\Cref{alg:singlepass}) is negligible, supporting the benefit of our approach.
  }
  \label{fig:accelerations}
\end{figure}


\textbf{Gradient momentum.}
In \citep{nesterov_momentum} it was shown that adding gradient momentum accelerates convergence substantially. We observed that this is also the case in our setting.

\textbf{Optimal initialization.}
From \Cref{app:convexity}, we see that the $\gL_2$ objective is minimized at the zero-gradient value, \ie $\mX_{opt} = \mM \pdot \measmat$.
Initializing $\mX$ like this already positions it in the feasible set, resulting in faster convergence and stabler behavior, particularly with momentum.

\textbf{Early termination.}
Optimal initialization leads to stable behavior of $\lVert \mX \rVert_*$, monotonically decreasing and then stabilizing.
This allows for a straightforward and fully automated way of identifying when to stop the optimization:
We run an exponential moving average gathering the change in $L_1$ loss from one iteration to the next one, and when it is close enough to zero, the optimizer terminates.
This makes the algorithm adaptive: if more deflation is needed, it will run for longer.

\textbf{Compact recovery.}
The \textsc{singlepass} recovery method from \citet{tropp_svd17} (\Cref{alg:singlepass}) is scalable since it avoids performing more expensive measurements of $\mA$.
However, it still requires to perform numerical routines with $\bigdim \!\times\! \measdim$  (\ie \emph{thin}) operators, which can be time-consuming for large $\bigdim$ \citep[\eg][]{fernandez}.
Here, we propose a \textsc{compact} recovery method (\Cref{alg:compact}) that relies for the most part on $\measdim \!\times\! \measdim$  (\ie \emph{compact}) operations.
Following \citet{rmt_svd}, we first QR-decompose our random measurements as $\mX\!\eqDef\!\mP \mS$ and $\mW\!\eqDef\!\mQ \mR$, such that $\mL \approx \mP \mP^* \mL \mQ \mQ^*$.
But, instead of discarding $\mS$, we incorporate it in the derivations as follows:
 \begin{align}
   \mP \mS \!=\! \underbrace{\mU \mSigma \mV^*}_{\mL} \measmat \approx \mU \mSigma \!\!\underbrace{\mV^* \mQ}_{\mZ^* \text{(unitary)}} \!\! \underbrace{\mQ^* \measmat}_{\mB}
   ~~~\Rightarrow
   \begin{cases}
     (\mS \mB^\dagger)^* (\mS \mB^\dagger) \!\approx\!\mZ \mSigma^2 \mZ^* ~~\Rightarrow \mV\!\approx\!\mQ \mZ\\
     \mP\!\approx\!\mU \mSigma \mZ \mB \mS^\dagger ~~~\Rightarrow \mU\!\approx\!\mP (\mSigma \mZ \mB \mS^\dagger)^*
   \end{cases}
 \end{align}
\\[-0.5em]
As a result, we see that both pseudoinverses are indeed \emph{compact}.
This recovery also features a $\measdim \!\times\! \measdim$ Hermitian eigendecomposition, instead of an $\bigdim \!\times\! \measdim$ SVD.
In exchange, this method requires two $\bigdim \!\times\! \measdim$ QR decompositions instead of one.
In terms of stability we note that, since $\mP$ and $\mQ$ are orthonormal, our pseudoinverses retain the same good condition as the ones from \citet{tropp_svd17}.
We empirically verify this, by comparing the recovery error $\rho^2$ and runtime for both single-pass and compact methods.
We observe a speedup, even at small scales, without any performance degradation (\Cref{fig:accelerations}).
This modest speedup is expected to grow at larger scales (see \citet[5.5]{golub13} for asymptotics).
Importantly, this proposed compact recovery is not \sketchlord-specific and can directly replace single-pass steps in other sketched algorithms (see \cref{sec:synth}).




\subsection{Stability and Versatility}
\label{sec:stability}

\AF{Note: in \cite[1.4]{rpca} it is shown closed-form solutions for $\lambda$, and a large range of stable ones.}

With the accelerations introduced in last section \sketchlord has 5 hyperparameters: step size and momentum for the gradient step ($\eta, \mu$), threshold for the shrinkage step ($\lambda$), $\mu$ and a decay/tolerance for the early termination (\ie the exponential moving average).
Here, we focus on ($\eta, \mu, \lambda$), studying their relation and showing that the behavior of \sketchlord is stable across many configurations.

\begin{table}[ht]
  \centering
  \caption{\textbf{Convergence results for different hyperparameters:} See \Cref{sec:stability} for discussion.}
  \label{tab:stability}
   \scriptsize
  \resizebox{0.85\textwidth}{!}{%
    \begin{tabular}{ c c S[table-format=4.1(3)] c S[table-format=1.2(3)e-1] S[table-format=1.2(3)e-1] c }
      \toprule
          {\textbf{$\eta$}} & {\textbf{$\lambda$}} &  {\textbf{$\mu$}} & {Iterations} & {\textbf{$\rho_{tot}^2$} ($\downarrow$)} & {\textbf{$\rho_{diag}^2$}  ($\downarrow$)} & {Convergence} \\
          \midrule
          1     & 0.05     & 0.95  & 492.4(52.7)      & 4.04(475)e-3  & 5.34(205)e-3  & \textbf{***} \\
          0.5   & 0.05     & 0.95  & 504.9(67.1)      & 3.20(463)e-2  & 2.88(235)e-2  & ** \\
          0.25  & 0.05     & 0.95  & 557.7(22.0)      & 7.42(1500)e-2 & 8.96(1030)e-2 & ** \\
          0.125 & 0.05     & 0.95  & 553.9(26.0)      & 1.10(134)e-1  & 2.09(641)e-1  & * \\
          \midrule
          1     & 0.2      & 0.95  & 582.7(36.2)      & 5.30(157)e-1  & 3.47(896)e-1  & * \\
          1     & 0.1      & 0.95  & 564.0(33.6)      & 1.22(274)e-1  & 9.45(199)e-2  & ** \\
          1     & 0.025    & 0.95  & 536.0(0.0)       & 5.03(119)e-3  & 4.07(804)e-3  & \textbf{***} \\
          1     & 0.0125   & 0.95  & 443.4(62.0)      & 6.32(148)e-4  & 6.07(86)e-4   & \textbf{****} \\
          \midrule
          0.5   & 0.025    & 0.95  & 522.4(26.1)      & 4.02(490)e-3  & 5.19(209)e-3  & \textbf{***} \\
          0.25  & 0.0125   & 0.95  & 462.0(77.5)      & 4.22(476)e-3  & 5.22(207)e-3  & \textbf{***} \\
          \midrule
          0.5   & 0.025    & 0.99  & 2240.1(294.0)    & 1.00(314)e-2  & 6.29(192)e-3  & ** \\
          0.25  & 0.0125   & 0.99  & 2201.9(328.9)    & 9.50(298)e-3  & 5.83(178)e-3  & ** \\
          \midrule
          0.5   & 0.025    & 0.5   & 1092.7(3.6)      & 4.73(973)     & 2.60(470)     & $\emptyset$ \\
          0.25  & 0.0125   & 0.5   & 2109.7(6.3)      & 4.70(966)     & 2.59(466)     & $\emptyset$ \\
          \bottomrule
    \end{tabular}
  }
\end{table}

For each hyperparameter configuration, we sample $10$ \texttt{exp(0.5)} matrices with $\bigdim \!=\!500$ (see \cref{sec:synth,fig:synth_matrices}), and gathered recovery errors and runtimes.
\Cref{tab:stability} presents the main results, grouped in 5 sub-experiments:
\emph{(a)} Starting from a default, untuned parametrization, we explore values for $\eta$, observing that performance decreases with $\eta$. Note, due to the quadratic nature of the gradient-based objective, $\eta \!<\! 2$, to prevent divergence.
\emph{(b)} Then, we explore different values for $\lambda$, observing that performance is inversely proportional.
\emph{(c)} We then scale $\eta$ \& $\lambda$ simultaneously, observing that performance remains virtually equal.
Finally, we repeat the experiments from group \emph{(c)} but with \emph{(d)} larger and \emph{(e)} smaller momentum. We observe that target performance is still invariant to $\sfrac{\lambda}{\eta}$, but extreme momentum values can drastically hinder performance.
Our main conclusion is that the algorithm works fairly well for a broad range of hyperparameters as long as $\sfrac{\lambda}{\eta}$ is large enough.

As we will see in the next section, this stability is paired with competitive performance on a broad class of matrices and great diagonal recovery (\Cref{app:synth_plots}).
The downside of extra computations related to the ADMM optimization is easily offset by the fact that measurements are now bounded by $\rank(L)$, not $\rank{A}$.
Furthermore, the early termination ties the amount of computation to deflation.




%
\section{Synthetic experiments}
\label{sec:synth}

To empirically validate our theoretical analysis (\cref{sec:toy}) and assess the practical performance of \sketchlord, we now conduct recovery experiments on an extensive suite of synthetic matrices.
These experiments are designed to evaluate \acf{LoRD} recovery, from approximate \ac{LoRD} but also purely low-rank matrices, and compare \sketchlord against established sketched baselines (see \cref{sec:toy_experiment}).
Full details of the experimental setup, including matrix generation and hyperparameter settings, are provided in \cref{app:synth}.

\textbf{Experimental setup.}
We synthetically generate approximately \ac{LoRD} matrices of the form $\mL + \xi \mD$, where $\mL$ is approximately low-rank and $\mD$ is a diagonal matrix. 
Following \citet[7.3.1]{ssvd19}, we construct $\mL$ using nine different matrix types, featuring three spectral decays (\texttt{exp}, \texttt{poly}, and \texttt{noise}) each with three different noise levels (denoted in parenthesis, \eg \texttt{exp(0.5)}).
See \cref{fig:synth_matrices} for an illustration of these decays and the resulting matrices, and \citet[7.3.1]{ssvd19} for details.
The diagonal component $\mD$ is sampled from a Gaussian distribution, and its relative prominence is controlled by a factor $\xi$, which scales the diagonal's norm relative to the norm of $\mL$.
We test $\xi \!\in\! \{0, 0.1, 1, 10\}$, representing purely low-rank, weak diagonal, balanced, and strong diagonal dominance, respectively.
For all cases, we evaluate three configurations, pairing matrix sizes $\bigdim \in \{500, 1000, 5000\}$ with corresponding approximate ranks $\smalldim \in \{10, 20, 100\}$ and the same measurement counts $p \in \{180, 360, 900\}$ for all methods, respectively.
Lastly, we check all methods with three different recovery strategies (single-pass, oversampled, and compact) and $30$ matrix samples per setting, resulting in more than $9000$ evaluated matrices.
\begin{figure}[t]
	\centering
	\includegraphics[width=0.95\textwidth]{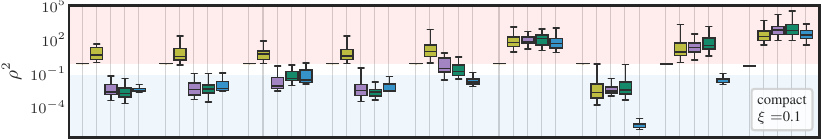}
	\includegraphics[width=0.95\textwidth]{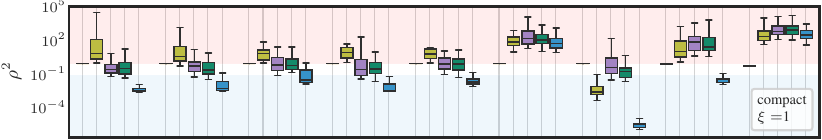}
	\hspace*{0.8mm}\includegraphics[width=0.95\textwidth]{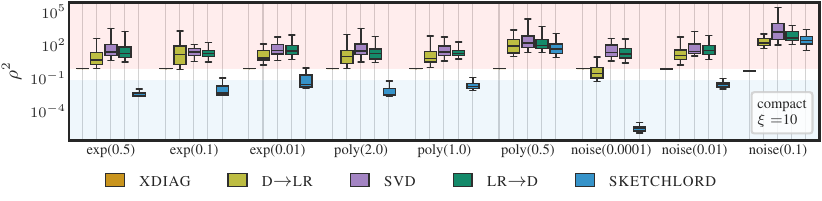}
	\caption{\textbf{\sketchlord provides high-fidelity approximations for \ac{LoRD} matrices.}
		Recovery results for various random \acf{LoRD} matrices of size $\bigdim\!=\!5000$. All matrices have an approximate rank $\smalldim\!=\!100$ with rank noise of different intensities (see x-axis and \Cref{app:synth} for more explanation). The relative strength of the diagonal D is given by $\xi$ and increases from the top to the bottom subplot.
		Provided are boxplots, across $30$ samples each, of the recovery error for our studied algorithms, all from $900$ measurements and \textsc{compact} recovery (\Cref{alg:compact}).
		Areas of $\rho^2 \geq100\%$ relative error are shaded in red,
		and $\rho^2 \leq10\%$ in blue.
	}
	\label{fig:main_results}
\end{figure}
All experiments are real-valued, and Rademacher random matrices are used for sketching, consistent with common practice.
Our baselines include \xdiag for purely diagonal recovery, \ssvd for purely low-rank approximations, and $\text{D} \!\rightarrow\! \text{LoR}$ (Diagonal-then-Low-Rank) as well as $\text{LoR} \!\rightarrow\! \text{D}$ (Low-Rank-then-Diagonal), with the implementation optimizations described in \cref{sec:toy_experiment}.
The approximation performance is primarily measured by the overall residual energy $\rho^2(\mA)$, \ie the relative approximation error in Frobenius norm (\cref{eq:rho}).
We also capture the residual energy of only the diagonal part, to measure the algorithms' ability to accurately estimate diagonals.

\textbf{Main results.}
Our synthetic experiments, summarized in \cref{fig:main_results}, with additional detailed plots in \cref{app:synth_plots}, reveal sever key insights into \sketchlord's performance.
\textbf{(I) Importance of joint recovery.}
As the relative strength of the diagonal component ($\xi$) increases, \sketchlord increasingly shows stronger recovery performance, and the benefits of a joint recovery become more apparent.
For matrices with a strong diagonal component, \eg $\xi \!=\!10$, \sketchlord consistently outperforms the baseline methods, often achieving recovery errors that are many order of magnitude lower.
In those cases, neither a purely diagonal approximation (\ie \xdiag) nor sequential approaches are sufficient to capture the composite structure.
This highlights the critical operational regime where \sketchlord should be the preferred method.
\textbf{(II) Efficiency of compact recovery.}
Furthermore, the compact recovery strategy we introduced in \cref{sec:sketchlord_algo} yields nearly identical results that the standard single-pass recovery, while offering improved computational efficiency.
\textbf{(III) Robustness in purely low-rank scenarios.}
For purely low-rank matrices ($\xi=0$), see \Cref{fig:lowrank_results}, \ssvd is the theoretically favored baseline and also empirically tends to yield the best results among the compared methods.
However in many purely low-rank matrices, \sketchlord maintains commendable performance, sometimes even outperforming \ssvd (\eg \texttt{noise(0.0001)}) in terms of approximation quality.
This resilience suggests \sketchlord's versatility as drop-in alternative, even in cases where the presence or strength of a diagonal component in an operator is unknown.

In essence, \sketchlord successfully addresses scenarios where conventional sketched methods falter.
For \ac{LoRD} matrices, particularly those with strong diagonal parts, it offers substantially more accurate approximations, thereby expanding the scope of high-fidelity structured matrix recovery.

\section{Conclusion}
\label{sec:conclusion}

This paper addressed the suboptimal recovery of \acf{LoRD} operators by conventional sketched methods.
We introduced \sketchlord, a sketched algorithm based on convex optimization and ADMM, for joint \ac{LoRD} estimation.
Experiments on synthetic data 
demonstrate \sketchlord's ability to achieve high-fidelity approximations of composite \ac{LoRD} structures.

Despite its strong performance, \sketchlord has limitations, some inherited from the nature of sketched methods, \eg that the optimal noise structure for \ac{LoRD} recovery is currently unknown.
However, the most significant current drawback is the computational runtime, primarily due to the \acs{ADMM} optimizer requiring a \ac{SVD} in each iteration.
We see several avenues to mitigate this.\FS{If we need to save space, we could move our suggestions to the appendix, \ie  ``We see several avenues to mitigate this, detailed in Appendix A.''}
Future work could explore additional \acs{ADMM} accelerations, such as incorporating quasi-Newton methods like L-BFGS \citep{Liu1989} for the gradient step or FISTA \citep{fista} for the overall iteration.
Additionally, optimizing directly on the singular space of the operator could allow pruning singular values as they approach zero during optimization, potentially reducing computational load and memory as optimization progresses.
Beyond efficiency, the core framework of \sketchlord might be extendable:
By vertically shifting the measurement operator $\bar{\mOmega}$, we could target off-diagonal bands, potentially generalizing our method to the broader class of low-rank plus banded linear operators.
\AF{Also: more research on hyperparameters, particularly $\lambda$}

Ultimately, \sketchlord enriches the available toolkit of structured approximation techniques.
While other methods may prioritize speed, \sketchlord distinguishes itself by delivering superior approximation quality within a manageable memory footprint.
It therefore offers a compelling option for scenarios where high fidelity is paramount, and the investment in runtime is justified by the significant gains in accuracy.
\FS{I have some alternative formulations for the last paragraph in the comments (thanks LLMs!)}

\newpage
\bibliographystyle{abbrvnat}
\bibliography{references}


\appendix
\newpage
\section{Appendix}
\label{sec:appendix}

%
\subsection{Additional algorithms}
\label{app:algorithms}

This section provides additional pseudocode and background for the sketched algorithms involved in our experiments, in addition to the novel ones already provided in \Cref{sec:sketchlord}.

The original version of the \textsc{ssvd} (for \emph{Sketched SVD}) algorithm was introduced in \citet{rmt_svd}.
This version featured a recovery step that requires a \emph{second round of measurements} from the linear operator $\mA$, and for this reason is not considered here.
Instead, we gather in \Cref{alg:ssvd} the version from \citet{tropp_svd17}, which features a \textsc{singlepass} recovery (\cref{alg:singlepass}) that \emph{does not require any extra measurements}.
This is also the case for our proposed \textsc{compact} recovery (\Cref{alg:compact}).

An \textsc{oversampled} recovery, gathered in \Cref{alg:oversampled}, was introduced in \citet{boutsidis}. This recovery requires \emph{extra measurements, but they can be fully parallelized}.
In exchange for the extra measurements, it was shown in \citet{ssvd19}---and is confirmed by our experiments---that it leads to better approximations, particularly in scenarios like streaming settings or slowly decaying spectra.
We note that \emph{any of the three recovery algorithms can be used exchangeably for any of the sketched methods}, and we do so in our experiments (see results in \Cref{app:synth_plots}).

Lastly, we have the algorithms discussed in \Cref{sec:toy_experiment}: \textsc{xdiag} (introduced in \citet{xtrace} and gathered in \Cref{alg:xdiag}), and our two \emph{sequential} baselines: $\text{LoR}\!\rightarrow\!\text{D}$ (\Cref{alg:lowrank_then_diag}) and  $\text{D}\!\rightarrow\!\text{LoR}$ (\Cref{alg:diag_then_lowrank}).

\begin{minipage}[t]{0.48\textwidth}
  {
\begin{algorithm}[H]
  \DontPrintSemicolon
  \KwIn{$\mA\!\in\!\mathbb{C}^{\bigdim\!\times\!\bigdim}$ \tcp*{Linear operator}}
  \KwIn{$\mOmega, \mUpsilon \!\in\!\mathbb{C}^{\bigdim\!\times\!\measdim}$   \hspace{-15pt} \tcp*{Random matrices}}
  \tcp{Random measurements}
  $\mM \leftarrow \mA \mOmega$ \hspace{-1cm} \tcp*{$\mathbb{C}^{\bigdim\times \measdim}$}
  $\mW^* \leftarrow \mUpsilon^* \mA$ \hspace{-1cm} \tcp*{$\mathbb{C}^{\measdim \times \bigdim}$}
  \tcp{Recovery ($\mU, \mV\!\in\!\mathbb{C}^{\bigdim\!\times\!\measdim}$)}
  $(\mU, \mSigma, \mV^*) \leftarrow \textsc{singlepass}(\mM, \mOmega, \mW)$
  \Return{$(\mU, \mSigma, \mV^*)$} \tcp*{$\mU \mSigma \mV^* \approx \mA$}
  \caption{\textsc{ssvd} \citep{tropp_svd17}}
  \label{alg:ssvd}
\end{algorithm}
}

  {
\begin{algorithm}[H]
  \DontPrintSemicolon
  \KwIn{$\mM, \mOmega, \mW \!\in\!\mathbb{C}^{\bigdim\!\times\!\measdim}$}
  $(\mP, \_) \leftarrow \qr(\mW)$  \tcp*{$\sC^{\bigdim \times \measdim}$}
  $\mB \leftarrow \mM (\mP^* \mOmega)^\dagger$  \tcp*{$\sC^{\bigdim \times \measdim}$}
  $(\mU, \mSigma, \mZ^*) \leftarrow \svd(\mB)$ \tcp*{$\sC^{\bigdim \times \measdim}, \sC^{\measdim \times \measdim}, \sC^{\measdim \times \measdim}$}
  $\mV \leftarrow \mP \mZ$ \tcp*{$\sC^{\bigdim \times \measdim}$}
  \Return{$(\mU, \mSigma, \mV^*)$}
\caption{\textsc{singlepass} \citep{tropp_svd17}}
\label{alg:singlepass}
\end{algorithm}
}

    {
\begin{algorithm}[H]
  \DontPrintSemicolon
  \KwIn{$\mA\!\in\!\mathbb{C}^{\bigdim\!\times\!\bigdim}$ \tcp*{Linear operator}}
  \KwIn{$\mOmega \!\in\!\mathbb{C}^{\bigdim\!\times\!\measdim}$   \hspace{-15pt} \tcp*{Rademacher matrix}}
  \tcp{Random measurements and XDiag matrix}
  $\mM \leftarrow \mA \mOmega$ \hspace{-1cm} \tcp*{$\mathbb{C}^{\bigdim\times \measdim}$}
  $(\mQ, \mR) \leftarrow \qr(\mM)$\\
  $\mS \leftarrow \norm(\mR^\dagger)$ \tcp*{column normalization}
  $\mPsi \leftarrow \mI\!-\!\sfrac{1}{\measdim} \mS \mS^*$ \tcp*{$\mathbb{C}^{\measdim \times \measdim}$}
  \tcp{Top-space diagonal estimation}

  $\mX^* \leftarrow \mS \mQ^* \mA$ \\
  $\vd \leftarrow (\mQ \pdot \mX^*) \ones$\\
  \tcp{Bottom-space diagonal estimation}
  \For{$i \in \{1, \tightDots \measdim \}$ }{
    $\vd \leftarrow \vd + \big[\sfrac{1}{\measdim}~\vomega_i \pdot (\vm_i - \mQ \mX^* \vomega_i) \big]$
  }
  \Return{$(\vd, \mQ)$} \tcp*{$\vd \approx \diag(\mA)$}
  \caption{\textsc{xdiag} \citep[2.1]{xtrace}}
  \label{alg:xdiag}
\end{algorithm}
}

\end{minipage}
\hfill
\begin{minipage}[t]{0.48\textwidth}
  {
\begin{algorithm}[H]
  \DontPrintSemicolon
  \KwIn{$\mA\!\in\!\mathbb{C}^{\bigdim\!\times\!\bigdim}$ \tcp*{Linear operator}}
  \KwIn{$\mOmega, \mUpsilon, \mGamma \!\in\!\mathbb{C}^{\bigdim\!\times\!\measdim}$   \hspace{-15pt} \tcp*{Rademacher matrices}}
  $(\mU, \mSigma, \mV^*) \leftarrow \textsc{ssvd}(\mA, \mOmega, \mUpsilon)$\\
  $\mB \leftarrow (\mA - \mU \mSigma \mV^*)$ \tcp*{Implicit, matrix-free}
  $\vd \leftarrow \sfrac{1}{\measdim}(\mGamma \pdot \mB \mGamma) \ones$\\
  \Return{$(\mU, \mSigma, \mV^*), \vd$} \tcp*{$\mU \mSigma \mV^* + \diag(\vd) \approx \mA$}
  \caption{$\text{LoR}\!\rightarrow\!\text{D}$}
  \label{alg:lowrank_then_diag}
\end{algorithm}
}

  {
\begin{algorithm}[H]
  \DontPrintSemicolon
  \KwIn{$\mA\!\in\!\mathbb{C}^{\bigdim\!\times\!\bigdim}$ \tcp*{Linear operator}}
  \KwIn{$\mOmega, \mUpsilon, \mGamma \!\in\!\mathbb{C}^{\bigdim\!\times\!\measdim}$   \hspace{-15pt} \tcp*{Rademacher matrices}}
  $(\vd, \mQ) \leftarrow \textsc{xdiag}(\mA, \mOmega)$ \\
  \tcp{asdf}
  $\mM \leftarrow \mA \mOmega - \diag(\vd) \mOmega$  \tcp*{Recycled from XDiag}
  $\mW^* \leftarrow \mUpsilon^* \mA - \mUpsilon^* \diag(\vd)$ \\
  $(\mU, \mSigma, \mV^*) \leftarrow \textsc{compact}(\mM, \mOmega \mW)$ \\
  \Return{$(\mU, \mSigma, \mV^*), \vd$} \tcp*{$\mU \mSigma \mV^* + \diag(\vd) \approx \mA$}
  \caption{$\text{D}\!\rightarrow\!\text{LoR}$ }
  \label{alg:diag_then_lowrank}
\end{algorithm}
}

  {
\begin{algorithm}[H]
  \DontPrintSemicolon
  \KwIn{$\mA\!\in\!\mathbb{C}^{\bigdim\!\times\!\bigdim}$ \tcp*{Linear operator}}
  \KwIn{$\mP, \mQ \!\in\!\mathbb{C}^{\bigdim\!\times\!\measdim}$ \tcp*{Orthogonalized sketches}}
  \KwIn{$\mOmega', \mUpsilon' \!\in\!\mathbb{C}^{\bigdim\!\times\!2\measdim}$ \tcp*{Random matrices}}
  \tcp{Perform core measurements and solve core matrix}
  $\mC \leftarrow \mUpsilon'^*\!\mA \mOmega'$ \hspace{-1cm} \tcp*{$\mathbb{C}^{2\measdim \times 2\measdim}$}
  $\mC \leftarrow (\mUpsilon'^* \mP)^\dagger \mC (\mQ^* \mOmega')^\dagger$ \tcp*{$\mathbb{C}^{\measdim \times \measdim}$}
  $(\mU', \mSigma, \mV'^*) \leftarrow \svd(\mC)$\\ 
  $\mU \leftarrow \mP \mU'$ \tcp*{$\mathbb{C}^{\bigdim \times \measdim}$ orthonormal}
  $\mV \leftarrow \mQ \mV'$ \tcp*{$\mathbb{C}^{\bigdim \times \measdim}$ orthonormal}
  \Return{$(\mU, \mSigma, \mV^*)$}
\caption{\textsc{oversampled} \citep{boutsidis,ssvd19}}
\label{alg:oversampled}
\end{algorithm}
}

\end{minipage}

%

\subsection{Mathematical details for \cref{sec:toy_theory}}
\label{app:toy_theory}

Here, we provide a more detailed derivation for the error of a \textbf{low-rank-then-diagonal} approximation.
In this approximation, we start with low-rank deflation followed by diagonal estimation of the residual, \ie $\hat{\mA} \!=\! \lowrankApprox{\mA}_\smalldim \!+\! \diag(\mA \!-\! \lowrankApprox{\mA}_\smalldim)$.
By definition, and based on the Fourier eigendecomposition from \cref{sec:toy_theory}, the residual has the form $\mA \!-\! \lowrankApprox{\mA}_{\smalldim} \!=\! \sum_{i=\smalldim}^{\bigdim - 1} \vf_i \vf_i^*$ ($\smalldim \geq 1$), with scalar diagonal in the form $\diag(\mA \!-\! \lowrankApprox{\mA}_{\smalldim}) \!=\! \frac{\bigdim - \smalldim}{\bigdim} \mI$.
Also recall that $\lVert \mA \rVert_\frob^2 \!=\! \bigdim (\bigdim + 3)$ and that the Fourier vectors $\vf_i$ form a unitary basis.
With this, we derive the following recovery error:
\begin{align*}
	\rho^2_{\text{LoR} \rightarrow \text{D}} \big(\mA \big)
	&= \frac{\lVert \mA - \lowrankApprox{\mA}_\smalldim - \diag(\mA - \lowrankApprox{\mA}_\smalldim)  \rVert_\frob^2}{\lVert \mA \rVert_\frob^2} 
	= \frac{\lVert (\sum_{i=\smalldim}^{\bigdim - 1} \vf_i \vf_i^*)  -  \frac{\bigdim - \smalldim}{\bigdim} \mI  \rVert_\frob^2}{\bigdim (\bigdim + 3)} \\
	&= \frac{1}{\bigdim (\bigdim + 3)} \Bigg( \sum_{i=\smalldim}^{\bigdim - 1} \sum_{j=\smalldim}^{\bigdim - 1} (\vf_i^*\vf_j)^2  + \frac{(\bigdim - \smalldim)^2}{\bigdim} - 2 \frac{\bigdim - \smalldim}{\bigdim} \sum_{i=\smalldim}^{\bigdim - 1} (\vf_i^* \vf_i) \Bigg)\\
	&= \frac{(\bigdim - \smalldim) - \frac{(\bigdim - \smalldim)^2}{\bigdim}}{\bigdim (\bigdim + 3)}
	= \frac{\sfrac{\smalldim}{\bigdim} - \sfrac{\smalldim^2}{\bigdim^2}}{\bigdim + 3}
\end{align*}

\subsection{Extended discussion on optimization strategy}
\label{app:convexity}






Recall the $\gL_2$ component from \Cref{eq:problem_lambda}, derived from the equality constraint in \Cref{eq:equality_constraint}:
\begin{align}
   \gL_2 = \frac{1}{2} \lVert  \big(\mM - \mX \pdot \measmatBar \big) \big( \mI - \frac{1}{\measdim}\ones \ones^* \big) \rVert_\frob^2
\end{align}
This can be equivalently expressed in vectorized form via Kronecker products as follows \citep[10.2.2]{cookbook}:
\begin{align}
  \gL_2 = &\frac{1}{2}\lVert \big( \underbrace{\mI - (\frac{1}{\measdim}\ones \ones^* \otimes \mI  )}_{\mPhi} \big) \big( \underbrace{\vecc(\mM)}_{\vm} -  \underbrace{\diag(\vecc(\measmatBar))}_{\bar{\mJ}} \underbrace{\vecc(\mX)}_{\vx} \big) \rVert_2^2
  = \frac{1}{2} \lVert \mPhi \vm - \mPhi \bar{\mJ} \vx \rVert_2^2
\end{align}
Minimizing this loss is a classic linear least squares problem that, not only is \emph{convex} \wrt $\vx$, but also admits a \emph{closed-form} expression for the gradient \citep[5.1]{cookbook} that is particularly simple due to symmetry and idempotence of $\mPhi$:
\begin{align}
  \nabla_{\vx} \gL_2 = \frac{1}{2} (2 \bar{\mJ}^* \underbrace{\mPhi^* \mPhi}_{= \mPhi} \bar{\mJ} \vx - 2 \bar{\mJ} \mPhi^* \mPhi \vm) = \bar{\mJ} \mPhi (\bar{\mJ} \vx - \vm)
\end{align}
Which can be in turn expressed in matrix form by inverting the vectorization step, yielding the gradient as presented in \Cref{eq:admm_gradient}:
\begin{align}
  \nabla_{\mX} \gL_2 = \measmatBar \pdot \vecc^{-1}\!\big(\mPhi (\bar{\mJ} \vx - \vm) \big)
  = \measmatBar \pdot \big( (\mX \pdot \measmatBar) - \mM   \big) \big( \mI - \frac{1}{\measdim}\ones \ones^* \big) \,.
\end{align}
This derivation also yields a closed-form optimal value for $\mX$ that guarantees a value of zero for $\gL_2$ and its gradient.
In \Cref{sec:sketchlord_algo}, we show that initializing the ADMM solver with this $\mX_{opt}$ value results in faster convergence and stabler behavior, particularly with momentum:
\begin{align}
  \mX_{opt} \!=\! \mM \pdot \measmat ~\Rightarrow~  \mX_{opt} \pdot \measmatBar \!=\! \mM  ~\Rightarrow~
  \begin{cases}
    &\gL_2 \!=\!0\\
    \nabla_{\mX}\!\!\!\!\!\!\! &\gL_2 \!=\!0
  \end{cases}
\end{align}

To conclude this section, we provide the derivation of $\mD_\natural$ from \Cref{eq:diag_recovery_deflated_meas}. Recall that $\mD = \diag(\vd)$ and $\vd \ones^* = \mM - \mX \pdot \measmatBar$. Then,
\begin{align}
  \vd = \frac{1}{\measdim} \vd \ones^* \ones = \frac{1}{\measdim} \big( \mM - \mX \pdot \measmatBar \big) \ones\,.
\end{align}

%
\subsection{Synthetic Experiment Details}
\label{app:synth}

\begin{figure}[!htb]  
  \SetKwBlock{Repeat}{repeat}{}  
  \centering
  \scalebox{.9}{
    \begin{minipage}{1\linewidth}  
      \newcommand{\algoMargin}{0em}
      \IncMargin{\algoMargin}
      \begin{algorithm}[H]
	\DontPrintSemicolon
	\For{$(\bigdim, \smalldim, \measdim)\in \{(500, 5, 90), (1000, 10, 180), (5000, 50, 900)\}$ \tcp*{Matrix size, rank, \#measurements} }{
	  \For{$\xi \in \{0, 0.1, 1, 10 \}$ \tcp*{Diagonal strength}}{
	    \For{$\gM \in \{$ \texttt{exp(0.5)}, \texttt{exp(0.1)}, \texttt{exp(0.01)}, \texttt{poly(2)}, \texttt{poly(1)}, \texttt{poly(0.5)}, \texttt{noise(0.0001)}, \texttt{noise(0.01)}, \texttt{noise(0.1)}  $\}$  \tcp*{Low-rank matrix distribution}}{
              \Repeat(~\num{30} times){}{
                $\mL \sim \gM(\bigdim, \smalldim)$ \tcp*{Low-rank component $\sC^{\bigdim \times \bigdim}$ of approximate rank $\smalldim$}
                $\vd \sim \mathcal{N}(0, \mI)$ \tcp*{Diagonal component $\sC^{\bigdim}$}
                $\mA \leftarrow \mL + \xi \cdot \frac{\sfrac{\lVert \mL \rVert_2}{\sqrt{\bigdim}}}{\lVert \vd \rVert_2} \diag(\vd)$\\ 
                \For{$\gS \in \{$ \textsc{ssvd}, \textsc{xdiag}, \textsc{sketchlord} $\text{D}\!\rightarrow\!\text{LoR}$, $\text{LoR}\!\rightarrow\!\text{D}$  $\}$}{
                  \For{$\gR \in \{$ \textsc{singlepass}, \textsc{compact}, \textsc{oversampled} $\}$}{
                    $\hat{\mA} \leftarrow \gS_{\gR}(\mA, \measdim)$ \tcp*{Sketched estimation of $\mA$}
                  }
                }
              }
	    }
          }
        }
        \caption{Overview of the synthetic experimental setup discussed in \Cref{sec:synth,app:synth}. Each recovery algorithm features different hyperparametrizations and random matrix samples, which can be found in \Cref{app:algorithms}. The step of gathering recovery metrics and runtimes is omitted here.}
        \label{alg:synth_exp}
      \end{algorithm}
      \DecMargin{\algoMargin}
    \end{minipage}
  }
\end{figure}

\begin{figure}[h]
  \centering
  \includegraphics[width=0.99\textwidth]{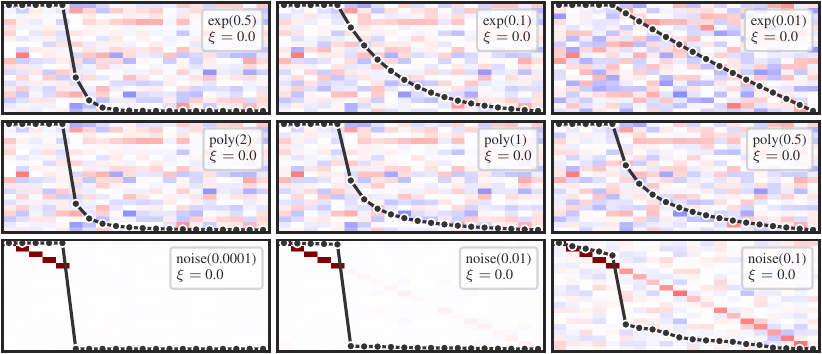}
  \\[1em]
  \includegraphics[width=0.99\textwidth]{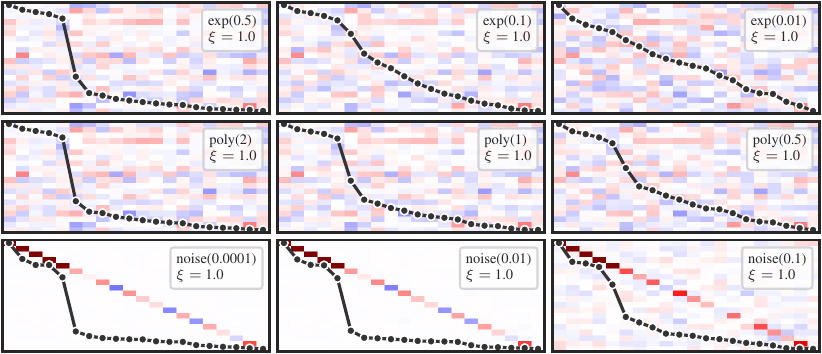}
  \\[1em]
  \includegraphics[width=0.99\textwidth]{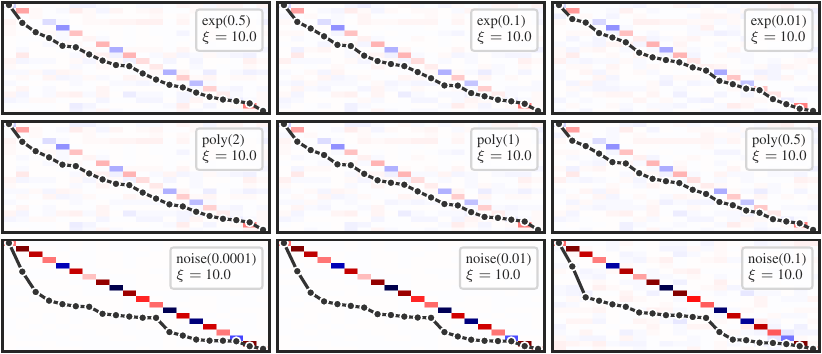}
  \caption{Different types of synthetic low-rank plus diagonal matrices, where $\xi$ expresses the \emph{relative importance} of the diagonal component.
     On top of each matrix, the corresponding distribution of singular values is provided, adjusted to fit the frame.
     See \Cref{app:synth} for more details.
  }
  \label{fig:synth_matrices}
\end{figure}

\textbf{Low-rank plus diagonal matrices:}
We follow \citet[7.3.1]{ssvd19} to sample three classes of \emph{approximately} low-rank random matrices $\mL$:
\emph{(a)} The \texttt{exp(t)} matrices feature $\smalldim$-many unit singular values, and then the following singular values decay exponentially, at a rate given by \texttt{t} (see \Cref{fig:synth_matrices}): as \texttt{t} increases, the effective rank gets closer to $\smalldim$. The left and right singular spaces are sampled randomly. Following \citet{ssvd19}, we use \texttt{t} values of \num{0.5} (fast decay), \num{0.1} (medium) and \num{0.01} (slow).
\emph{(b)} The \texttt{poly(t)} matrices are generated analogously to \texttt{exp(t)}, with the difference that, after the $\smalldim$ singular values of 1, the following ones decay polynomially instead of exponentially. Following \citet{ssvd19}, we use \texttt{t} values of \num{2} (fast decay), \num{1} (medium) and \num{0.5} (slow).
\emph{(c)} The \texttt{noise(t)} matrices are the sum of 2 matrices: a low-rank matrix that just has $\smalldim$-many unit diagonal entries (the rest is zeros), plus a noisy matrix with a noise intensity of \texttt{t}.
Following \citet{ssvd19}, we use \texttt{t} values of \num{0.0001} (low noise), \num{0.01} (medium) and \num{0.1} (high).

In order to make the sampled matrices low-rank \emph{plus diagonal} (\ie $\mL + \mD$), and still retain some control about the prominence of $\mD$, we introduce a \emph{diagonal ratio} parameter $\xi \in \R_{\geq 0}$: If $\xi\!=\!0$, no diagonal should be added. A large $\xi$ (\eg 10) should mean that $\mD$ is very prominent, and for $\xi\!=\!1$ it should be balanced.
This is realized by the following ratio between their norms:
\begin{align}
  \label{eq:diag_ratio}
  \mA = \mL + \xi \cdot \sqrt{ \frac{\lVert \mL \rVert_2^2}{\bigdim \lVert \mD \rVert_2^2} } \mD
\end{align}
We can see that, for $\xi\!=\!1$, the squared norm of the added diagonal, which has $\bigdim$ entries, equals the average squared norm of a vector from $\mL$, which also has $\bigdim$ entries.
See \Cref{fig:synth_matrices} for an illustration of this process and how it affects the appearance of the combined matrix, as well as its singular value distribution.

\textbf{Sketched mechanism:}
We note that, although our derivation of the sketch in \Cref{sec:sketchlord_convex} applies to any complex-valued matrix, our experiments are performed in $\mathbb{R}$.
In this case, the natural choice for random $\measmat, \measmatBar \!\in\! \sC^{\bigdim \!\times\! \measdim}$ satisfying $\measmat_{i, j} \measmatBar_{i, j} \!=\! 1 \,\forall\, (i, j)$ is to use Rademacher noise, since this is satisfied in an efficient and numerically stable manner for $\measmat = \measmatBar$.
While it would be interesting to explore other sources of noise, like SSRFT \citep[3.2]{ssvd19}, it has been noted that the particular source of noise does not typically make much difference in practice \citep[3.9]{tropp_svd17}.
Furthermore, Rademacher is a popular and well-studied source of sketching noise, including \citep{ssvd19,xtrace}, which helps establishing comparisons.

\textbf{Design choices:}
An overview of the experimental setup can be found in \Cref{alg:synth_exp}.
We sample square (but not symmetric) \ac{LoRD} matrices of 3 different sizes $\bigdim$, up to $\bigdim\!=\!5000$.
In order to decide the underlying rank $\smalldim$ and number of sketched measurements $\measdim$, we found that $\smalldim\!=\!\sfrac{\bigdim}{100}$ and $\measdim\!=\!18\smalldim$ exposed the regime in which algorithms would sometimes fail, and sometimes succeed at recovering the matrices (see \Cref{app:synth_plots}).
We chose $\measdim$ to be a multiple of 6 in order to ensure that all compared algorithms perform the exact same number of measurements (since some measure in groups of 2, and others in groups of 3).
As previously explained, we sample low-rank matrices from 9 different classes following \citet[7.3.1]{ssvd19}, and add diagonal components following 4 ratios $\xi$: \num{0} (no added diagonal), \num{0.1} (weak), \num{1} (balanced) and \num{10} (strong).
For each scenario, we sample 30 different random matrices, and provide error bars gathered across this population.
And for each sample, we ran all sketched algorithms with all 3 possible recovery mechanisms, gathering runtimes and performance metrics.

\textbf{Implementation:}
Performing the full experiment for each sample took $\sim$18 hours on a distributed CPU cluster (\ie $\sim$\num{540} hours total).


%
\subsection{Synthetic Experiment, Extra Results}
\label{app:synth_plots}

\begin{figure}[t]
  \centering
  \includegraphics[width=0.95\textwidth]{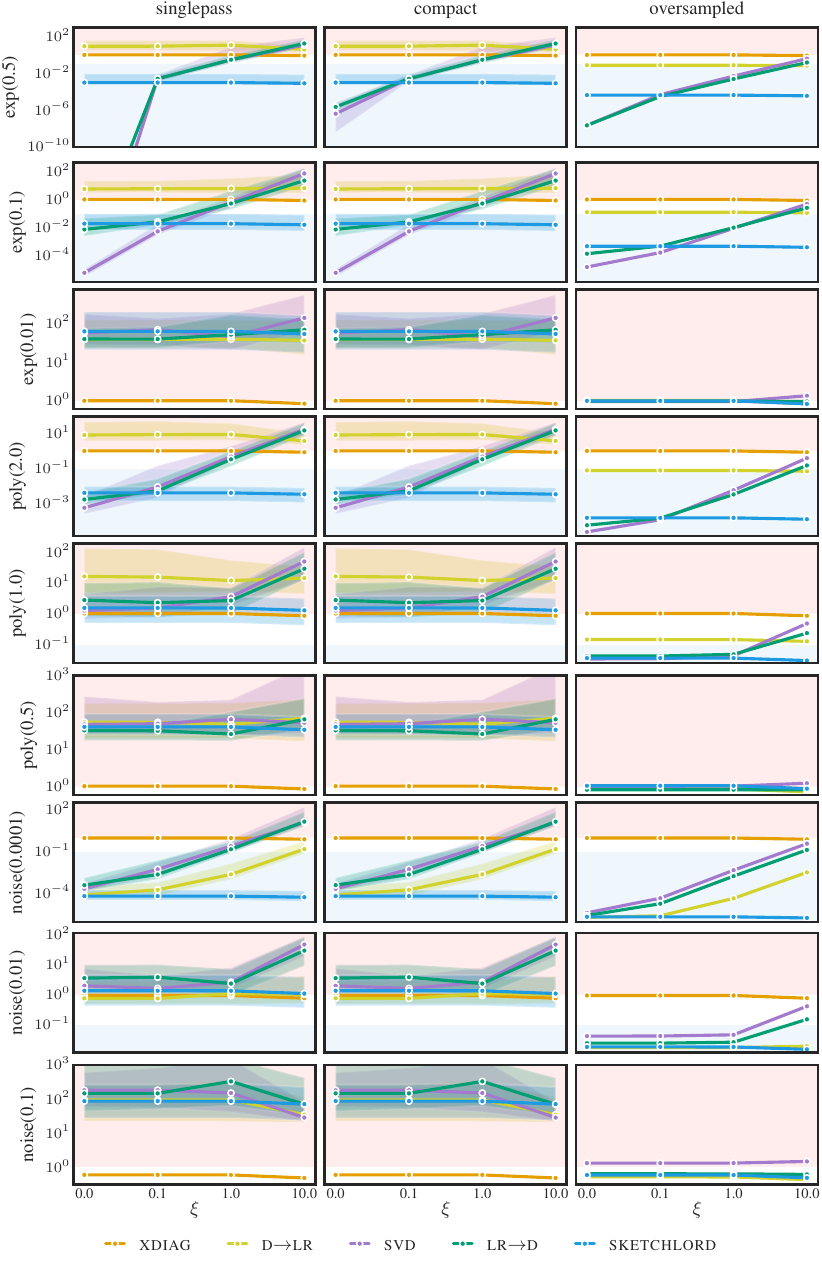}
  \caption{Residual energy ($\rho^2$) for the \textbf{full recovery of $500\times500$ matrices}, gathered following the protocol described in \Cref{app:synth}. Areas of $100\%$ error and above are shaded in red. Areas of $10\%$ error and below are shaded in blue. See \Cref{app:synth,app:synth_plots} for further details and discussion.
  }
  \label{fig:full_total_500}
\end{figure}

\begin{figure}[t]
  \centering
  \includegraphics[width=0.95\textwidth]{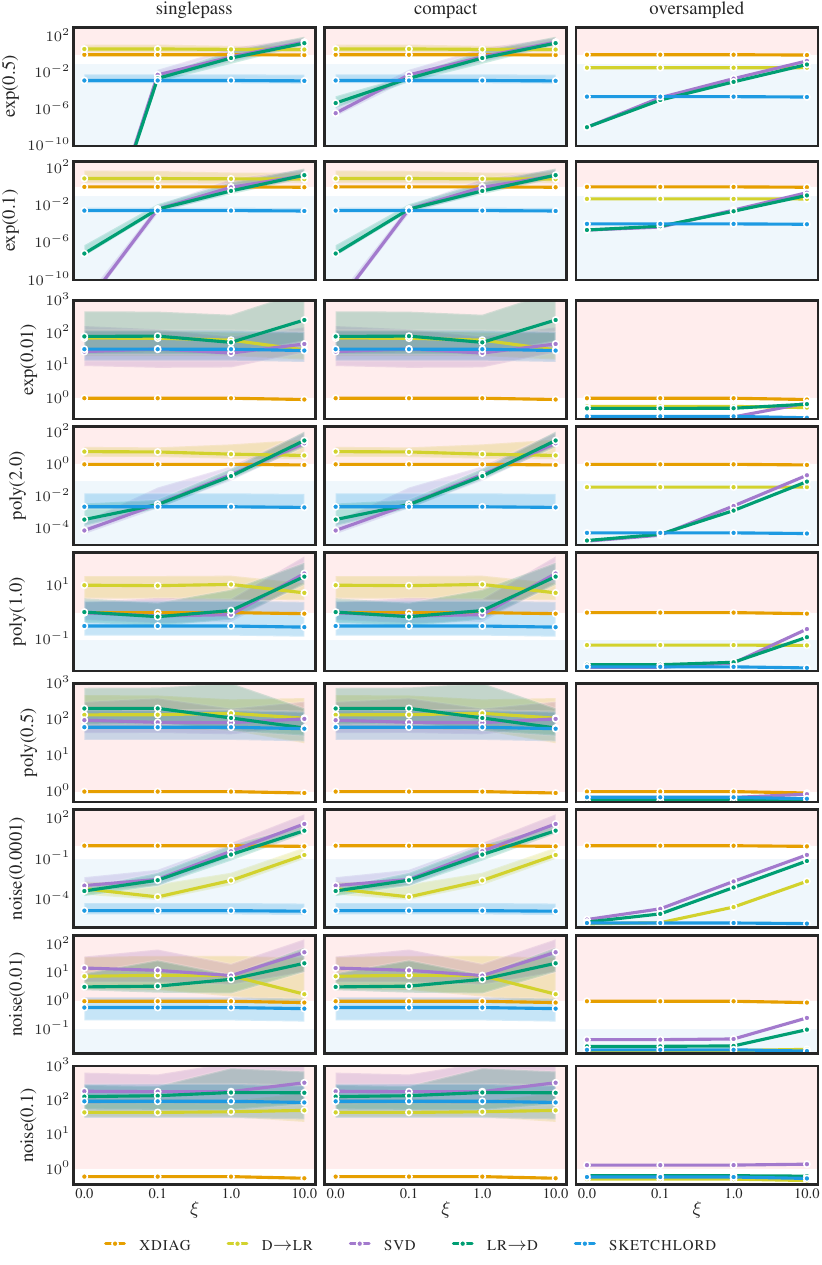}
  \caption{Residual energy ($\rho^2$) for the \textbf{full recovery of $1000\times1000$ matrices}, gathered following the protocol described in \Cref{app:synth}. Areas of $100\%$ error and above are shaded in red. Areas of $10\%$ error and below are shaded in blue. See \Cref{app:synth,app:synth_plots} for further details and discussion.
    }
  \label{fig:full_total_1000}
\end{figure}

\begin{figure}[t]
  \centering
  \includegraphics[width=0.95\textwidth]{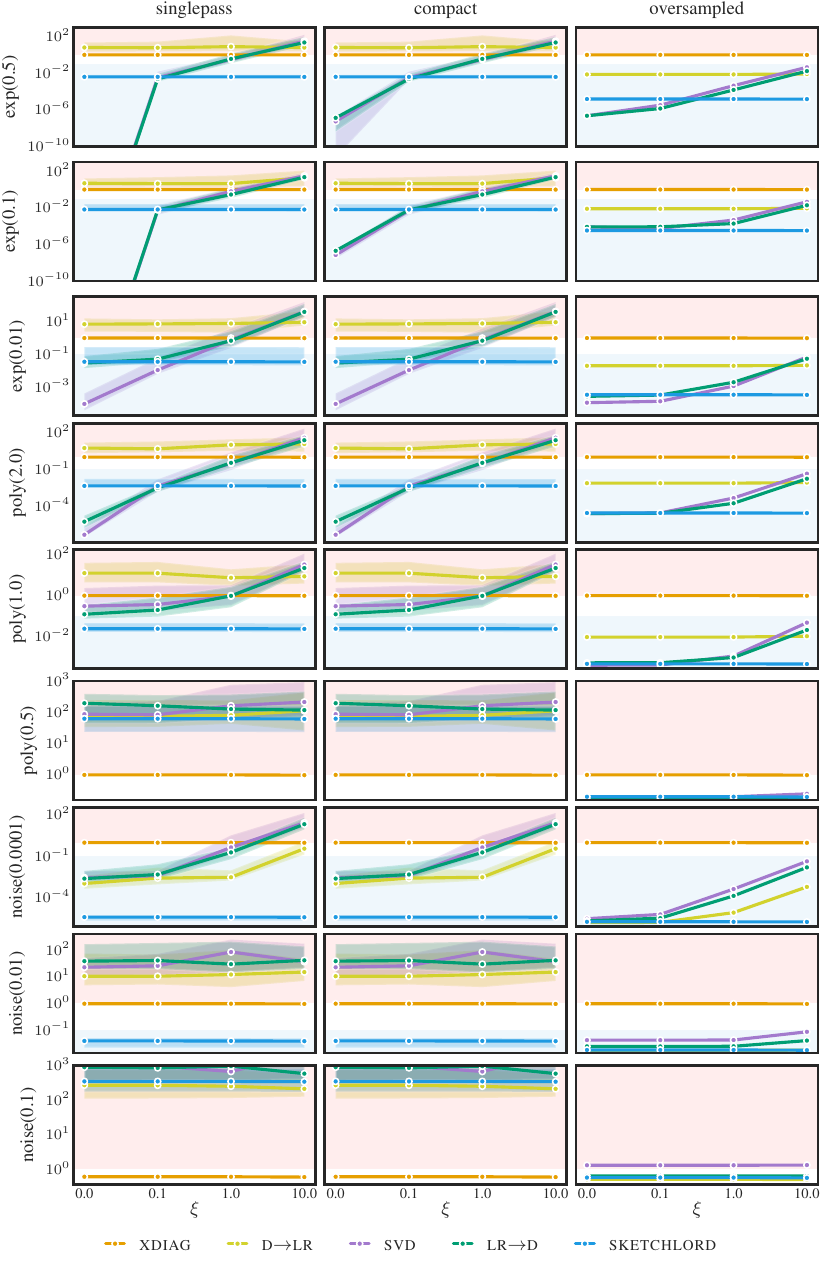}
  \caption{Residual energy ($\rho^2$) for the \textbf{full recovery of $5000\times5000$ matrices}, gathered following the protocol described in \Cref{app:synth}. Areas of $100\%$ error and above are shaded in red. Areas of $10\%$ error and below are shaded in blue. See \Cref{app:synth,app:synth_plots} for further details and discussion.
    }
  \label{fig:full_total_5000}
\end{figure}

\begin{figure}[t]
  \centering
  \includegraphics[width=0.95\textwidth]{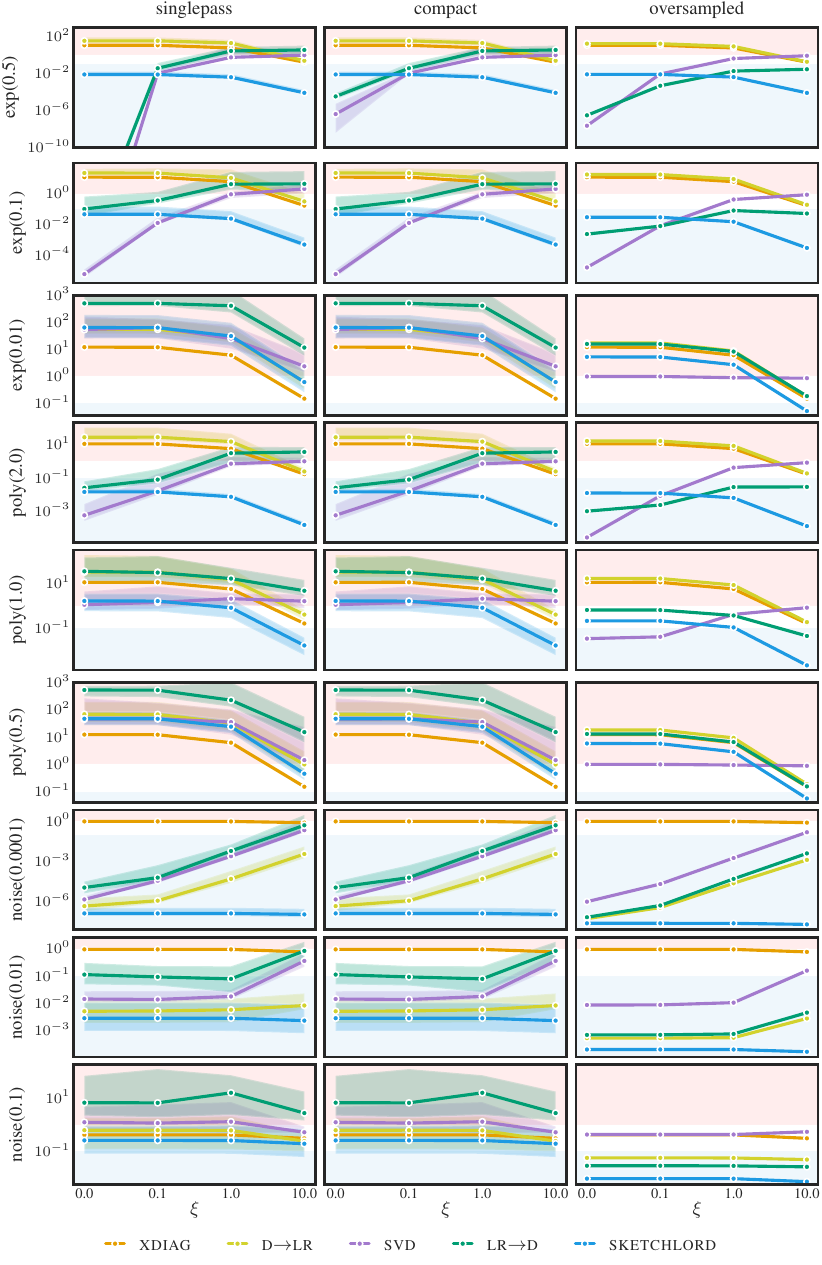}
  \caption{Residual energy ($\rho^2$) for the \textbf{diagonal recovery of $500\times500$ matrices}, gathered following the protocol described in \Cref{app:synth}. Areas of $100\%$ error and above are shaded in red. Areas of $10\%$ error and below are shaded in blue. See \Cref{app:synth,app:synth_plots} for further details and discussion.
  }
  \label{fig:full_diag_500}
\end{figure}

\begin{figure}[t]
  \centering
  \includegraphics[width=0.95\textwidth]{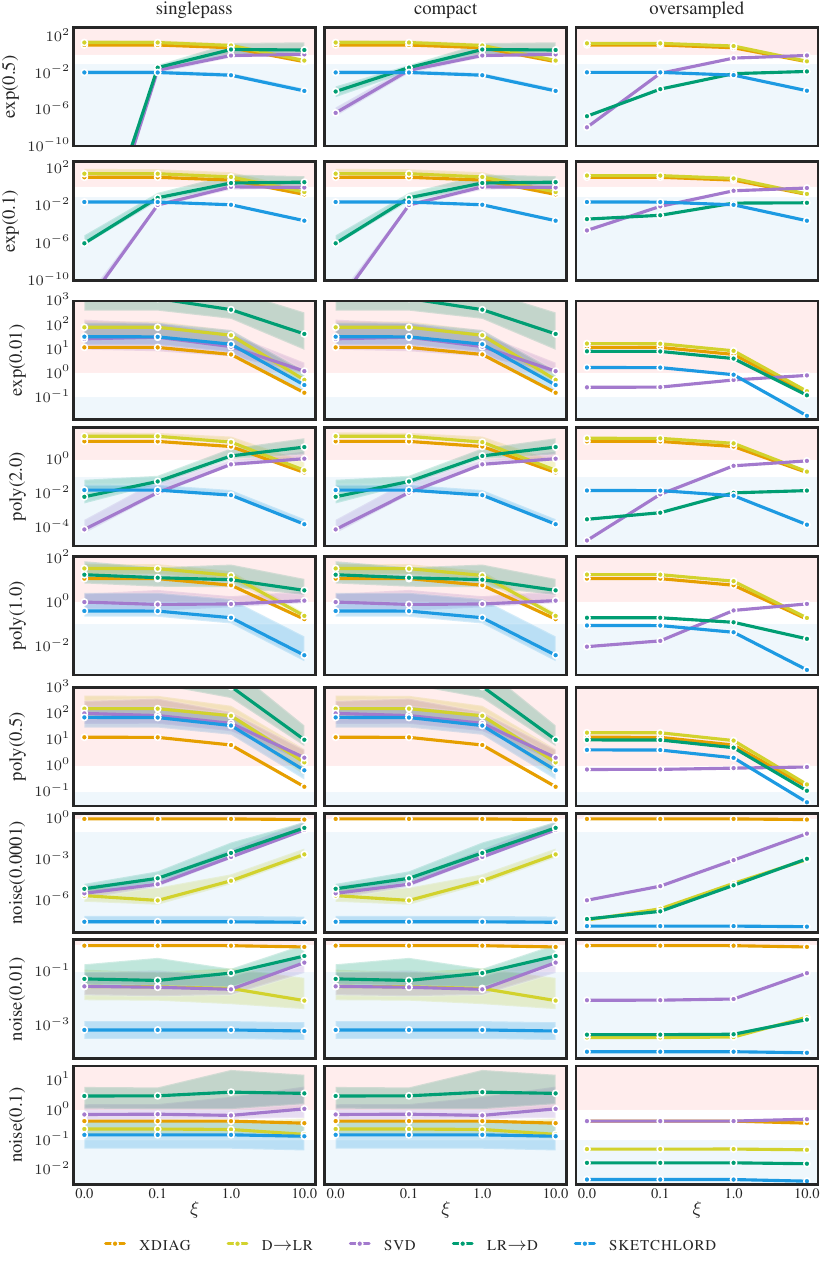}
  \caption{Residual energy ($\rho^2$) for the \textbf{diagonal recovery of $1000\times1000$ matrices}, gathered following the protocol described in \Cref{app:synth}. Areas of $100\%$ error and above are shaded in red. Areas of $10\%$ error and below are shaded in blue. See \Cref{app:synth,app:synth_plots} for further details and discussion.
  }
  \label{fig:full_diag_1000}
\end{figure}

\begin{figure}[t]
  \centering
  \includegraphics[width=0.95\textwidth]{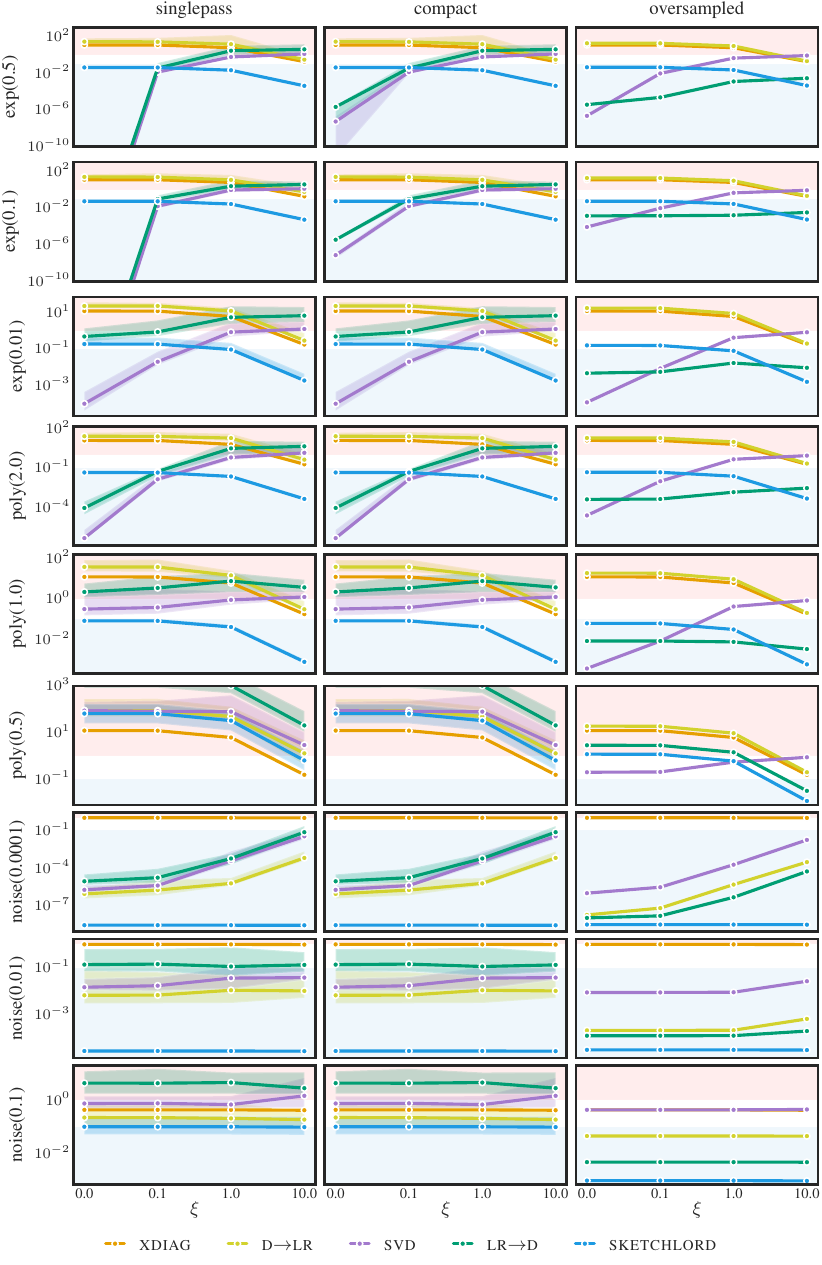}
  \caption{Residual energy ($\rho^2$) for the \textbf{diagonal recovery of $5000\times5000$ matrices}, gathered following the protocol described in \Cref{app:synth}. Areas of $100\%$ error and above are shaded in red. Areas of $10\%$ error and below are shaded in blue. See \Cref{app:synth,app:synth_plots} for further details and discussion.
    }
  \label{fig:full_diag_5000}
\end{figure}

\begin{figure}[t]
	\centering
	\includegraphics[width=0.95\textwidth]{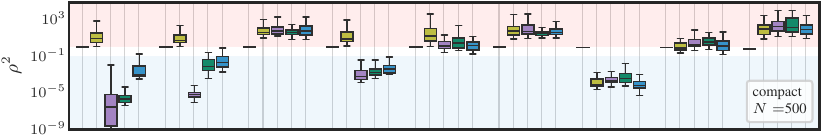}
	\includegraphics[width=0.95\textwidth]{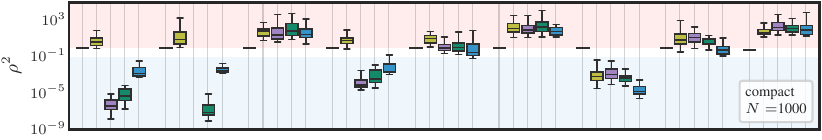}
	\hspace*{0.8mm}\includegraphics[width=0.95\textwidth]{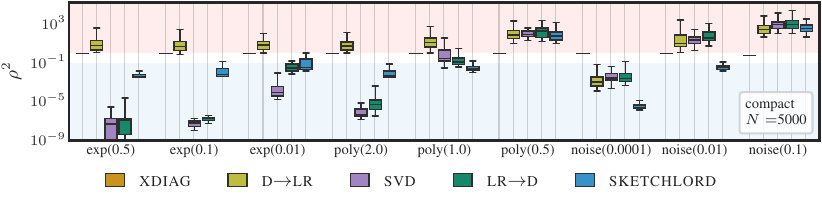}
	\caption{\textbf{\sketchlord is competitive for low-rank matrices, although generally not better than \textsc{ssvd}:} Recovery error for various matrix types and sizes at $\xi\!=\!0$ (\ie only the low-rank component is present). See \Cref{app:synth,app:synth_plots} for further details and discussion.}
	\label{fig:lowrank_results}
\end{figure}

\begin{figure}[t]
	\centering
	\includegraphics[width=0.95\textwidth]{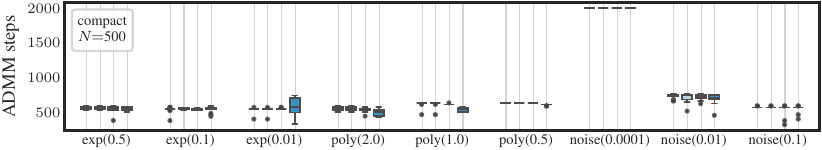}
        \includegraphics[width=0.95\textwidth]{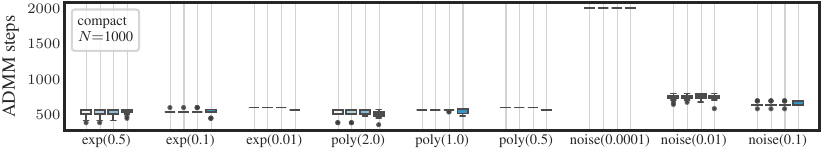}
	\hspace*{0.8mm}\includegraphics[width=0.95\textwidth]{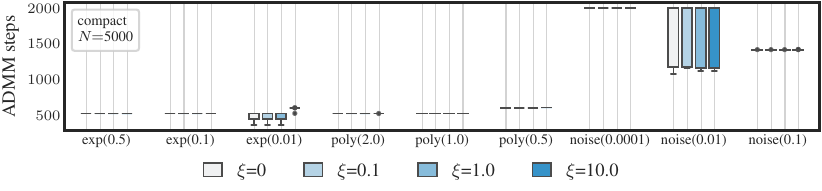}
	\caption{\textbf{Number of ADMM iterations in \textsc{sketchlord}}, for different matrix types and sizes. See \Cref{app:synth_plots} and \Cref{fig:runtime_results} for further discussion.
	}
	\label{fig:iter_results}
\end{figure}

\begin{figure}[t]
	\centering
	\includegraphics[width=0.952\textwidth]{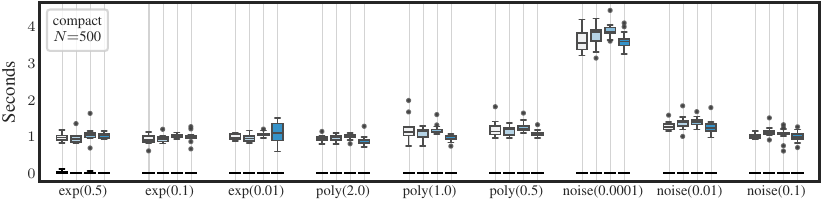}
        \hspace*{-3.5mm}\includegraphics[width=0.978\textwidth]{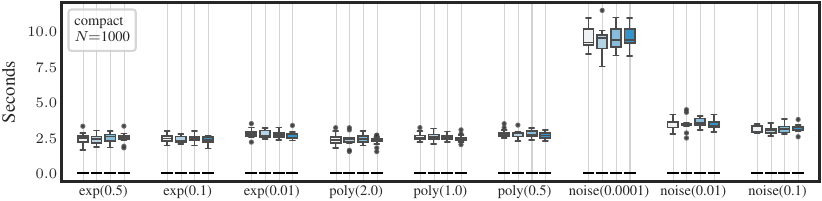}
	\hspace*{-2mm}\includegraphics[width=0.97\textwidth]{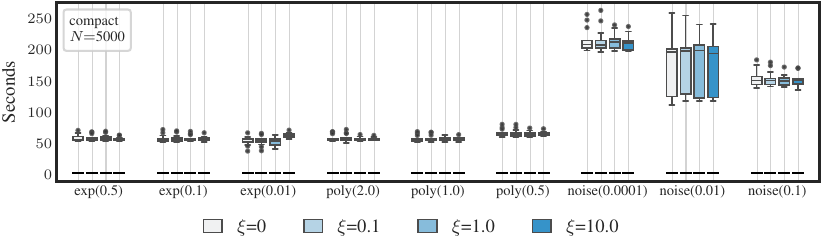}
	\caption{\textbf{Runtime of \textsc{sketchlord} \versus \textsc{ssvd}:} Total elapsed time, in seconds, for different matrix sizes and values of $\xi$. The thin boxes in the near-zero mark correspond to the \textsc{ssvd} runtimes, and the boxes above to \textsc{sketchlord}. Note how the overhead of \textsc{sketchlord} is mostly determined by the number of ADMM iterations (\Cref{alg:sketchlord} and \Cref{fig:iter_results}).}
	\label{fig:runtime_results}
\end{figure}

This section contains additional results from our experiments with synthetic matrices, extending \Cref{sec:synth} and \Cref{app:synth}:

\textbf{Full recovery line plots (\Cref{fig:full_total_500,fig:full_total_1000,fig:full_total_5000})}: These plots display the $\rho^2$ error between original and recovery, for all matrix types, algorithms, recoveries and diagonal ratios $\xi$. See \Cref{eq:rho,eq:diag_ratio} for the definitions of $\rho^2$ and $\xi$, respectively. Note how, when one method fails, all methods tend to also fail (likely due to insufficient measurements). Given sufficient measurements, \textsc{sketchlord} provides almost always the best recovery.\\
\textbf{Diagonal recovery line plots (\Cref{fig:full_diag_500,fig:full_diag_1000,fig:full_diag_5000}):} Same as above, but the $\rho^2$ error is measured between the \emph{diagonal} of the original and the \emph{diagonal} of the recovery. Thus, it measures the effectiveness in recovering the diagonal of a linear operator, even for cases when the operator itself is not diagonal. Note how, given enough measurements, the joint recovery strategy followed by \textsc{sketchlord} also yields superior results, even compared to ad-hoc methods like \textsc{xdiag}.\\
\textbf{Low-rank recovery boxplots (\Cref{fig:lowrank_results}):} In \Cref{fig:main_results}, the recovery error for various matrix types is provided, for cases where the diagonal component is nonzero. Here, we provide similar results but with $\xi\!=\!0$, \ie for the cases where the additional diagonal component is zero and just the low-rank component is present. Here, \textsc{sketchlord} is also competitive, although generally not better than \textsc{ssvd}.\\
\textbf{Runtime boxplots (\Cref{fig:iter_results,fig:runtime_results}):} \textsc{sketchlord} features an ADMM optimization step that can take a variable number of steps until termination, and thus have an impact in the overall runtime. These plots provide the gathered number of iterations and runtimes across all matrix types, sizes and diagonal ratios. Interestingly, we observe that the ADMM runtime does not seem to depend on $\xi$: this may be due to excessively conservative values for the early termination scheme. Furthermore, ADMM seems to run longer for matrices of the \texttt{noise} type, even when the matrix is simple and the final recovery is good. This may also indicate early termination issues for those cases.

%

\end{document}